\documentclass[10pt,twocolumn,letterpaper]{article}

\usepackage{cvpr}              

\newcommand{\red}[1]{{\color{red}#1}}

\newcommand{\simf}{\mathrm{sim}}

\usepackage{dsfont}
\usepackage{bm}
\usepackage{multirow}
\usepackage{adjustbox}
\usepackage{amsmath}
\usepackage{float}
\usepackage{caption}
\usepackage{amssymb}
\usepackage{subcaption}
\usepackage{verbatim}
\usepackage{multicol}
\usepackage{booktabs}
\usepackage{ulem}
\usepackage{comment}
\usepackage{soul,color} 
\usepackage{kotex} 
\usepackage{pifont}
\usepackage{multirow}
\usepackage{pifont}

\usepackage[linesnumbered,ruled]{algorithm2e}

\newcommand{\grey}[1]{\textcolor{gray!40}{#1}}
\usepackage{colortbl}
\usepackage{xcolor}
\definecolor{rowgray}{RGB}{240,240,240}
\definecolor{red}{RGB}{255, 102, 102}

\definecolor{cvprblue}{rgb}{0.21,0.49,0.74}
\usepackage[pagebackref,breaklinks,colorlinks,allcolors=cvprblue]{hyperref}

\def\paperID{16826} 
\def\confName{CVPR}
\def\confYear{2026}

\title{Learning to Refuse: Refusal-Aware Reinforcement Fine-Tuning for Hard-Irrelevant Queries in Video Temporal Grounding}

\author{Jin-Seop Lee$^{1}$\thanks{Equal contribution}\qquad SungJoon Lee$^{1}$\footnotemark[1]\qquad SeongJun Jung$^{1}$\qquad Boyang Li$^{2}$\qquad Jee-Hyong Lee$^{1}$\thanks{Corresponding author}\\
$^{1}$Sungkyunkwan University, South Korea\qquad
$^{2}$Nanyang Technological University, Singapore\\
{\tt\small \{wlstjq0602, sjoon8379, jsj412, john\}@skku.edu}\\ \tt\small boyang.li@ntu.edu.sg
}

\begin{document}
\maketitle
\begin{abstract}
Video Temporal Grounding (VTG) aims to localize a temporal segment in a video corresponding to a natural language query. 
However, existing VTG models assume that a relevant segment always exists, causing them to always predict a target segment even when the query is irrelevant to the video. 
While recent approaches attempt to handle irrelevant queries, they can only reject those that are entirely unrelated to the video and still fail to handle hard-irrelevant queries that are semantically similar but not actually relevant.
To address this, we propose Refusal-Aware Reinforcement Fine-Tuning (RA-RFT) to effectively refuse hard-irrelevant queries in VTG.
Our method is based on the Group Relative Policy Optimization (GRPO) framework and integrates four reward objectives—format, refuse-IoU, explain, and query correction—to improve both relevance discrimination and fine-grained semantic reasoning.
In addition, to effectively support RA-RFT, we construct a Hard-Irrelevant VTG (HI-VTG) dataset, which includes hard-irrelevant queries and their refusal answers.
We demonstrate the effectiveness of our method across various relevance-aware VTG scenarios, including hard-irrelevant VTG, simply-shuffled RA-VTG, and human-annotated RA-VTG settings. We also show that the proposed method is scalable by applying it to various LVLM-based VTG models. Our code is available at \url{https://github.com/JINSUBY/RA-RFT}.


\end{abstract}    
\section{Introduction}


Grounding target segments within a video for a user’s query is essential for real-world applications including video understanding agents and interactive video analysis systems~\cite{yang2025egolife, grauman2022ego4d, sigurdsson2016hollywood}. With this importance, research on Video Temporal Grounding (VTG), which aims to automatically extract relevant segments from videos based on a natural language query, has been actively explored in recent years~\cite{zhang2023temporal,tag,gao2017tall,mu2024snag, jang2023knowing}.
To understand time-sensitive semantics, VTG models are trained using pairs of natural language queries and their corresponding video moments.
Although they have achieved strong performances in temporal grounding, they rely on a strong assumption -- that a relevant segment always exists within the video.
As a result, most VTG models~\cite{ratsg,navmr} always predict a target segment even when the query is entirely unrelated to the video.

\begin{figure}[t]
\centering
\includegraphics[width=0.95\linewidth]{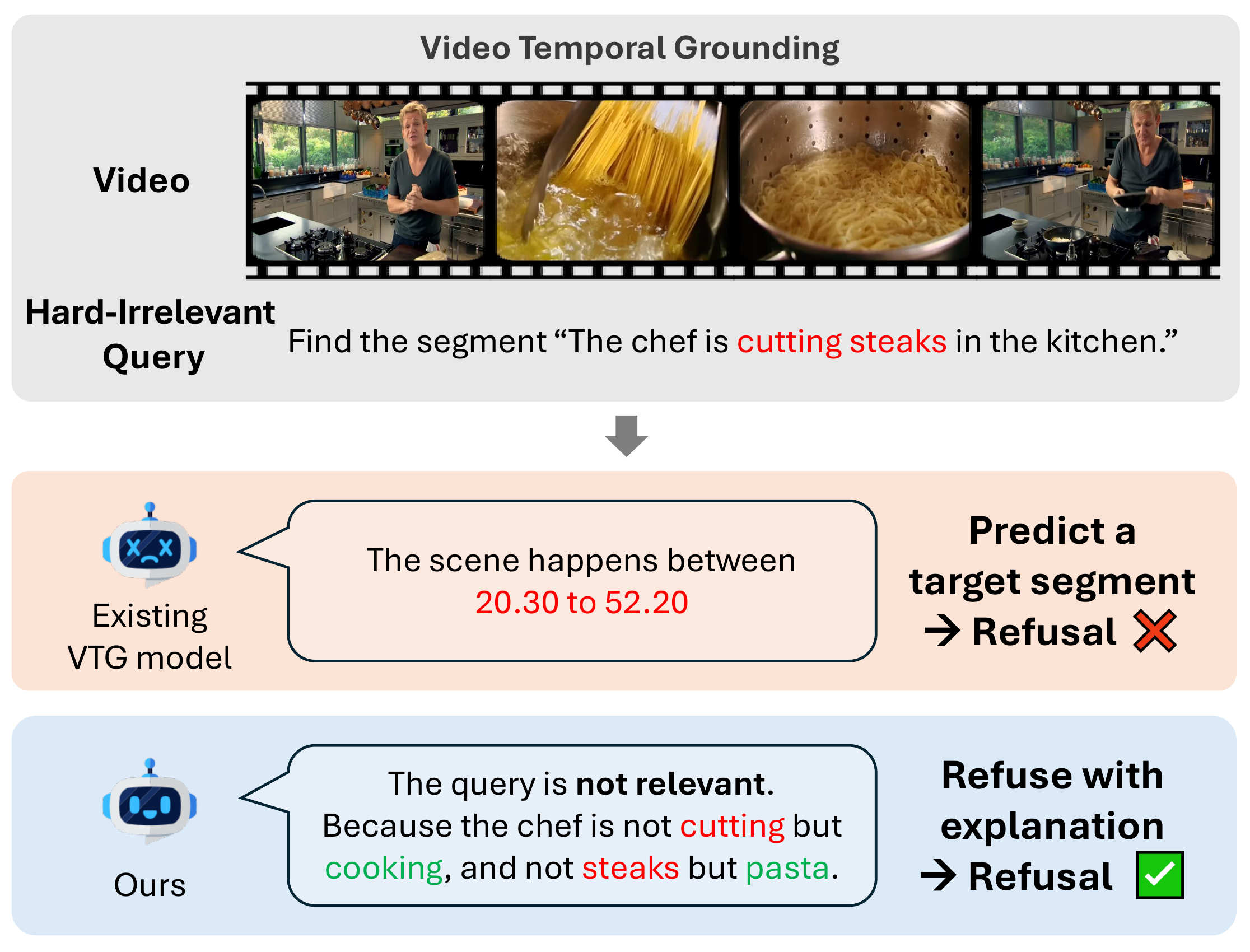}
\caption{
Video temporal grounding result with a hard-irrelevant query. Existing VTG models incorrectly predict a segment due to a lack of fine-grained semantic understanding between the video and the query. In contrast, our model correctly refuses the query and explains the semantic mismatch.
}
\label{fig_1}
\end{figure}

For considering real-world scenarios, the VTG model should accurately predict relevant segments when the query is relevant, while refusing to predict any segment when the query is irrelevant to the video. 
There are two possible ways to achieve this: training the model to explicitly refuse irrelevant queries, or leveraging the generalization capability of large vision-language models (LVLMs).
For training-based approaches~\cite{ratsg,navmr}, they construct irrelevant video–query pairs by simply shuffling queries across videos, and extract visual and textual features separately. Then, they train a binary classification layer to determine whether a given query is relevant to the video. 
For LVLM-based VTG approaches~\cite{time-r1,videochat-r1}, 
refusal-aware instructions can be added to the prompt to prevent segment prediction for irrelevant queries.
These approaches demonstrated good refusal performances on simply-shuffled irrelevant queries.
Despite these advances, they still have a limitation: they can only reject queries that are completely unrelated to the video, while failing on hard-irrelevant queries that are semantically close to the video but not actually relevant.

\Cref{fig_1} illustrates examples where a video and a hard-irrelevant query are given as input to a VTG model. 
The text query “The chef is cutting steaks in the kitchen” is semantically related to the video “The chef is cooking the pasta in the kitchen,” as both describe a chef preparing food in the kitchen. However, they differ in terms of the action (“cutting” vs, “cooking”) and the object (“steaks” vs. “pasta”), making the query not relevant to the video.
Nevertheless, existing approaches still fail to refuse these hard-irrelevant queries and predict target segments.
This happens because existing approaches cannot capture the fine-grained semantic differences between the video and the query.
Since training-based approaches are trained to refuse only completely unrelated queries (e.g., “the person is riding a bicycle”)~\cite{ratsg,navmr}, they primarily learn coarse-grained semantic differences and cannot distinguish fine-grained ones.
Also, LVLM-based VTG approaches are known to rely on coarse, high-level semantic cues, and thus cannot accurately capture fine-grained differences between the video and the query~\cite{hong2025motionbench, nguyen2024video}.
Therefore, to understand fine-grained semantic differences between the query and the video and to effectively refuse hard-irrelevant queries, it is necessary to develop a new learning strategy and construct a dataset that includes hard-irrelevant query–video pairs.

In this paper, we propose Refusal-Aware Reinforcement Fine-Tuning (RA-RFT) to effectively refuse hard-irrelevant queries in video temporal grounding.
We post-train a pretrained LVLM-based VTG model using Group Relative Policy Optimization (GRPO) and incorporate four reward objectives: a format reward, a refuse-IoU reward, an explain reward, and a query correction reward.
The refuse-IoU reward discourages segment prediction for hard-irrelevant queries and encourages accurate grounding for relevant ones.
The explain reward encourages the model to clearly explain why the query does not correspond to the video in irrelevant-query cases.
The query correction reward encourages reconstructing the relevant query from the given hard-irrelevant query and video context, enhancing the model’s reasoning ability for fine-grained semantic understanding.
These reward objectives not only improve relevance discrimination but also strengthen the model’s reasoning ability to understand fine-grained semantic differences between the query and the video. This strategy allows the model to make more accurate relevance judgments in hard-irrelevant cases.

In addition, to effectively support the proposed RA-RFT strategy, we construct a Hard-Irrelevant VTG (HI-VTG) dataset, which includes hard-irrelevant queries and refusal answers.
To generate hard-irrelevant queries, we extract relevance category types from the original queries and modify the original queries based on the extracted categories using an LLM. Then, we generate refusal answers using the video descriptions, original queries, irrelevant queries, and extracted relevance category types.
The HI-VTG training dataset consists of 2.5K relevant and 7.5K irrelevant query–answer pairs. We post-train the model on this dataset using our RA-RFT method.

We demonstrate the effectiveness of our method across various relevance-aware VTG scenarios, including hard-irrelevant VTG, simply-shuffled RA-VTG, and human-annotated RA-VTG settings. For all scenarios, our method improves not only relevance discrimination but also refusal explanation performance while maintaining temporal grounding performance. We also show that the proposed method can be applied to different LVLM-based VTG models~\cite{time-r1, videochat-r1}, demonstrating its scalability.

\label{sec:intro}

\begin{figure*}[t]
\centering
\includegraphics[width=1.0\linewidth]{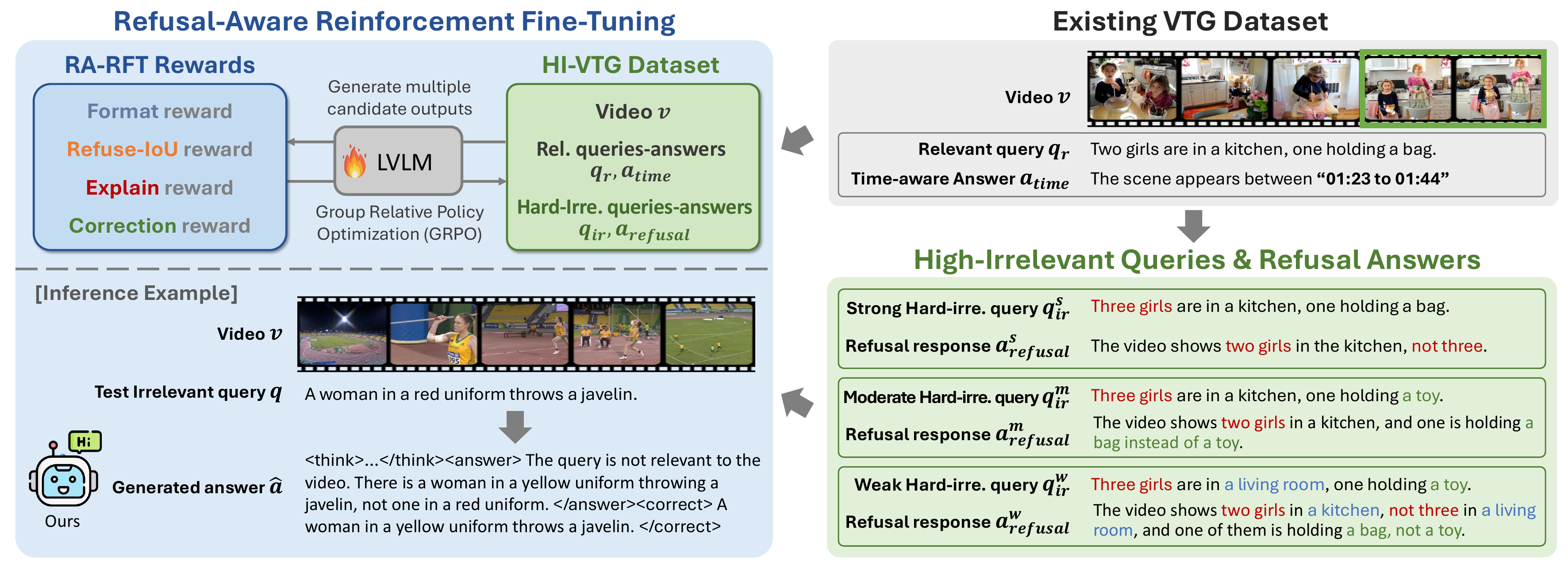}
\vspace{-0.5cm}
\caption{The overall framework of our contributions. We introduce a Hard-Irrelevant VTG Dataset, which includes hard-irrelevant queries and their refusal answers. Also, we propose a Refusal-Aware Reinforcement Fine-Tuning to effectively refuse hard-irrelevant queries.}
\label{fig_2}
\end{figure*}

\section{Related Work}
\subsection{Refusal-Capable Video Temporal Grounding}

Video Temporal Grounding (VTG) aims to automatically extract relevant segments from a video and a natural language query.
Most existing VTG models assume that a relevant segment always exists within a given video, so they predict a target segment even when the query is entirely unrelated to the video~\cite{zhang2023temporal,tag,gao2017tall,mu2024snag, jang2023knowing}. 
To address this problem, a few recent studies have explored VTG models capable of refusing predictions when the given query is irrelevant to the video~\cite{ratsg,navmr}.
RaTSG~\cite{ratsg} proposed a multi-task learning framework that jointly trains a relevance detection module and a temporal grounding module. NA-VMR~\cite{navmr} introduced an additional prediction head to determine the relevance of a given query.
However, since they are trained to refuse only completely unrelated queries, they primarily learn coarse-grained semantic differences and fail to capture fine-grained ones, making it difficult to refuse hard-irrelevant queries.
\vspace{0.3cm}
\subsection{Large Vision-Language Model-based VTG}
Traditional VTG methods~\cite{mu2024snag, jang2023knowing,ratsg, navmr} adopt a feature-fusion approach.
They extracted visual and textual features separately and fuse them to construct multi-modal representations. Then, they trained a target segment prediction model based on the fused multi-modal representations.
However, these models often suffer from limited generalization across visual and textual modalities, resulting in poor performance on unseen videos.

To address these issues, recent studies have explored LVLM-based VTG approaches, which build upon LLMs with strong generalization and reasoning capabilities.
These approaches are categorized into supervised fine-tuning (SFT)-based and reinforcement fine-tuning (RFT)-based VTG methods.
SFT-based VTG models~\cite{timechat, trace, timesuite,chatvtg} reformat existing video–timestamp pairs into an instruction-tuning format, and fine-tune the pretrained LVLM to enhance temporal grounding.
RFT-based VTG models~\cite{timezero,time-r1,videochat-r1} introduce time-aware reward functions under the GRPO framework, which has demonstrated strong reasoning improvements in LLMs such as DeepSeek~\cite{deepseek-r1,deepseekmatch}, to enhance temporal reasoning in videos.

Despite these successes, existing LVLM-based VTG models still fail to refuse hard-irrelevant queries, even when the prompt explicitly instructs the model not to output a target segment for irrelevant queries.
SFT-based VTG models are trained to imitate instruction-formatted answers and suffer from catastrophic forgetting of generalization capabilities~\cite{mitigates, shenfeld2025rl}, leading them to predict target segments regardless of query relevance.
RFT-based VTG models can refuse completely unrelated queries but still struggle with hard-irrelevant ones, as they rely on coarse, high-level semantic cues and fail to capture fine-grained semantic differences between the video and the query~\cite{hong2025motionbench, nguyen2024video}.
Therefore, a new training strategy is required to enable LVLM-based VTG models to understand fine-grained semantic differences and effectively refuse hard-irrelevant queries.

\section{Proposed Method}

\subsection{Preliminaries}
\label{sec_3_1}
\paragraph{Overall Framework.}
\label{sec_3_1_1}

To effectively refuse hard-irrelevant queries, we introduce a Refusal-Aware Reinforcement Fine-Tuning (RA-RFT) in video temporal grounding.
Also, we propose a Hard-Irrelevant VTG (HI-VTG) dataset to support the RA-RFT strategy.

\Cref{fig_2} illustrates the overall framework of the proposed RA-RFT strategy and HI-VTG dataset.
The existing VTG dataset consists of a video $v$, a relevant query $q_r$, and a time-aware answer $a_{\text{time}}$ that specifies the start and end timestamps of the target segment.
Based on this dataset, we generate a hard-irrelevant query $q_{ir}$ and its corresponding refusal answer $a_{\text{refusal}}$ using an LLM, where $a_{\text{refusal}}$ explains why the query does not correspond to the video.
Then, we construct the HI-VTG training dataset in the form $\{v, q_{r}, a_{\text{time}}, q_{ir}, a_{\text{refusal}}\}$, where each video is paired with both a relevant query and its time-aware answer, and a hard-irrelevant query with its refusal explanation.

Using this dataset, we post-train the VTG model via Group Relative Policy Optimization (GRPO), incorporating four reward objectives: a format reward, a refuse-IoU reward, an explain reward, and a query correction reward.
These reward objectives are designed to not only improve relevance discrimination but also enhance the model’s reasoning ability to understand fine-grained semantic differences between the query and the video, allowing the model to make more accurate relevance judgments in hard-irrelevant cases.

\paragraph{Background of GRPO.}
\label{sec_3_1_2}
Group Relative Policy Optimization (GRPO)~\cite{deepseek-r1, deepseekmatch} is a reinforcement learning algorithm that enhances the reasoning capability of LLMs and LVLMs through a predefined reward function~\cite{feng2025video,zhao2025r1,zhang2025r1,zhou2025r1,zhan2025vision,liu2025seg,deng2025boosting,liu2025visual,peng2025lmm,deng2025openvlthinker,yang2025r1}.
Given an input question $q$, the policy model $\pi_\theta$ generates a group of $G$ candidate responses $o = \{o_1, \dots, o_G\}$. Then, the reward function r(·) assigns a reward score to each response, yielding ${r(o_1), . . . , r(o_G)}$.
GRPO normalizes the reward scores by computing their mean and standard deviation, and encourages the model to generate responses that maximize a weighted-sum reward 
$R(o)$, defined as:
\begin{equation}
R(o) = \sum_i^G
\frac{\pi_\theta(o_i)}{\pi_{\theta_{\text{old}}}(o_i)}
\cdot
\frac{r(o_i) - \text{mean}(\{r(o_j)\}_{j=1}^{G})}
{\text{std}(\{r(o_j)\}_{j=1}^{G})}.
\label{eq:grpo_reward}
\end{equation}
where $\pi_{\theta}(o)$ represents the likelihood of the LLM generating a response $o$, and $\pi_{\theta_{\text{old}}}$ denotes the model parameters from the previous optimization step. To maintain training stability and avoid excessive deviation from the original model behavior, we incorporate a KL-divergence regularization term that penalizes the divergence between $\pi_{\theta}$ and the reference policy $\pi_{\text{ref}}$.
The final training objective is defined as follows:
\begin{equation}
\max_{\pi_\theta}
\mathbb{E}_{o \sim \pi_{\theta_{\text{old}}}(p)}
\Big[\, R(o) - \beta D_{\mathrm{KL}}(\pi_\theta \| \pi_{\text{ref}}) \Big],
\label{eq:grpo_objective}
\end{equation}
where $\beta$ controls the regularization strength, preventing excessive deviation from the reference model $\pi_{\text{ref}}$.

\subsection{Refusal-Aware Reinforcement Fine-Tuning}
\label{sec_3_2}
We aim to refuse hard-irrelevant queries and enhance the model’s reasoning ability to capture fine-grained semantic differences between the query and the video.
To achieve this, we propose Refusal-Aware Reinforcement Fine-Tuning (RA-RFT) based on GRPO.
Given a video $v$, relevant query-answer pairs $\{q_r, a_{\text{time}}\}$, and irrelevant query-answer pairs $\{q_{ir}, a_{\text{refusal}}\}$, we design four reward objectives for RA-RFT.
The overall reward, consisting of format, refuse-IoU, explain, and query correction rewards, is defined as:
\begin{equation}
r(o) = r_{\text{for}}(o) + r_{\text{R-IoU}}(o) + r_{\text{exp}}(o) + r_{\text{cor}}(o)
\label{eq:total_reward}
\end{equation}

\paragraph{Format Reward.}
The format reward $r_{\text{for}}(\cdot)$ encourages the LVLM to generate the output $o$ in the predefined template
{\small``\texttt{<think>...</think> <answer>...</answer> <correct>...</correct>}"}.
In this format, the \texttt{<think>} section describes the reasoning process to understand the temporal context of the video, the \texttt{<answer>} section either predicts the target segment or provides a refusal response, and the \texttt{<correct>} section reconstructs the relevant query when the input query is hard-irrelevant.
This reward encourages the model to organize its reasoning and prediction steps in a consistent structure.


\begin{equation}
r_{\text{for}}(o) =
\begin{cases}
1, & \text{if $o$ has correct format,}\\
0, & \text{if $o$ has wrong format.}
\end{cases}
\label{eq:format_reward}
\end{equation}




\paragraph{Refuse-IoU Reward.}
The VTG model should accurately predict relevant segments when the query is relevant, while refusing to predict any segment when the query is irrelevant to the video.
To handle both relevant and hard-irrelevant queries, we define the refuse-IoU reward as:
\begin{equation}
r_{\text{R-IoU}}(o) =
\begin{cases}
\text{IoU}(a_{time}, \hat{a}), 
& q \in \text{Rel.} \ \text{and} \ \{t_s, t_e\} \in \hat{a},\\[4pt]
1, 
& q \in \text{Irre.} \ \text{and} \ \{t_s, t_e\} \notin \hat{a},\\[4pt]
0, 
& \text{otherwise}.
\end{cases}
\label{eq:iou_reward}
\end{equation}
where $\hat{a}$ denotes the answer extracted from the generated output $o$ between \texttt{<answer>} and \texttt{</answer>}.
If the generated answer $\hat{a}$ contains a valid timestamp segment, we denote it as $\{t_s, t_e\} \in \hat{a}$; otherwise, $\{t_s, t_e\} \notin \hat{a}$.
For relevant queries, the reward assigns the IoU score to encourage accurate temporal localization when the model outputs a valid temporal segment.
For hard-irrelevant queries, the reward assigns a score of 1 only when the model does not output any temporal segment, encouraging refusal behavior.

\begin{figure*}[t]
\centering
\includegraphics[width=1\linewidth]{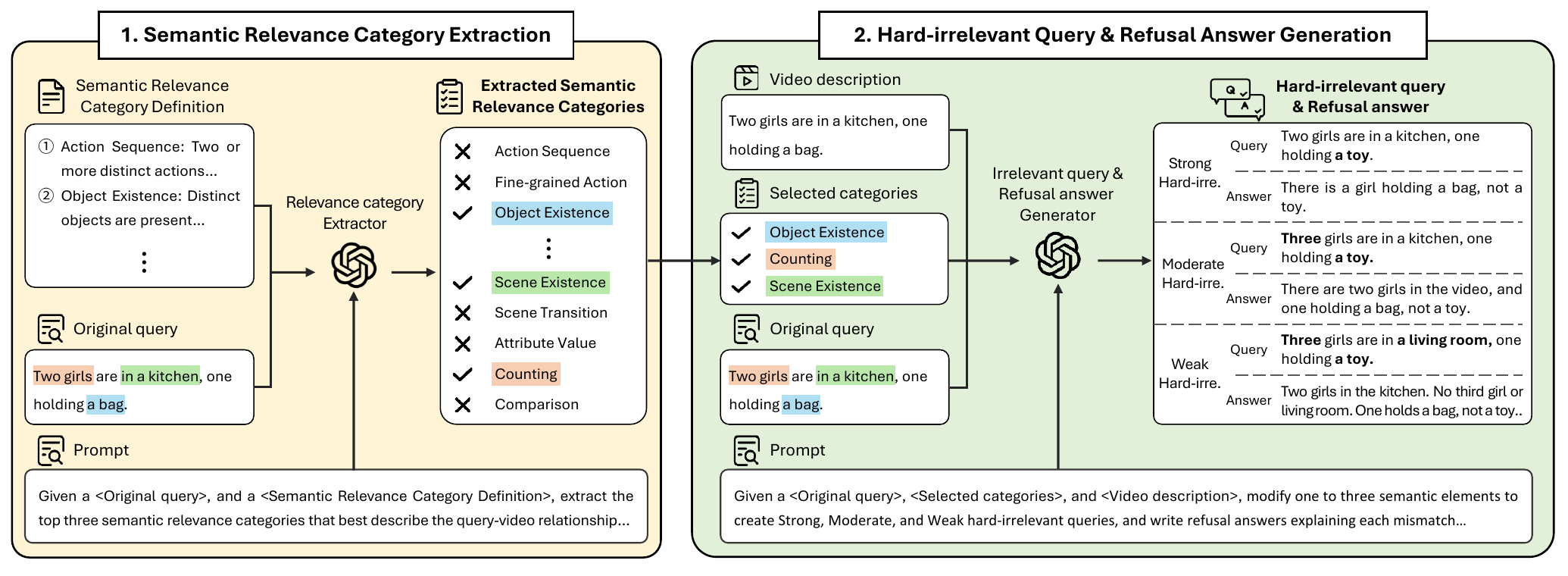}
\vspace{-0.5cm}
\caption{Overview of the Hard-Irrelevant VTG dataset construction process.
(1) We first extract semantic relevance categories from the original query using an LLM-based category extractor.
(2) Based on the selected categories and the video description, we then generate a hard-irrelevant query and its corresponding refusal answer, which explains why the query does not match the video.}
\label{fig_3}
\vspace{-0.2cm}
\end{figure*}

\begin{figure}[t]
\centering
\includegraphics[width=1\linewidth]{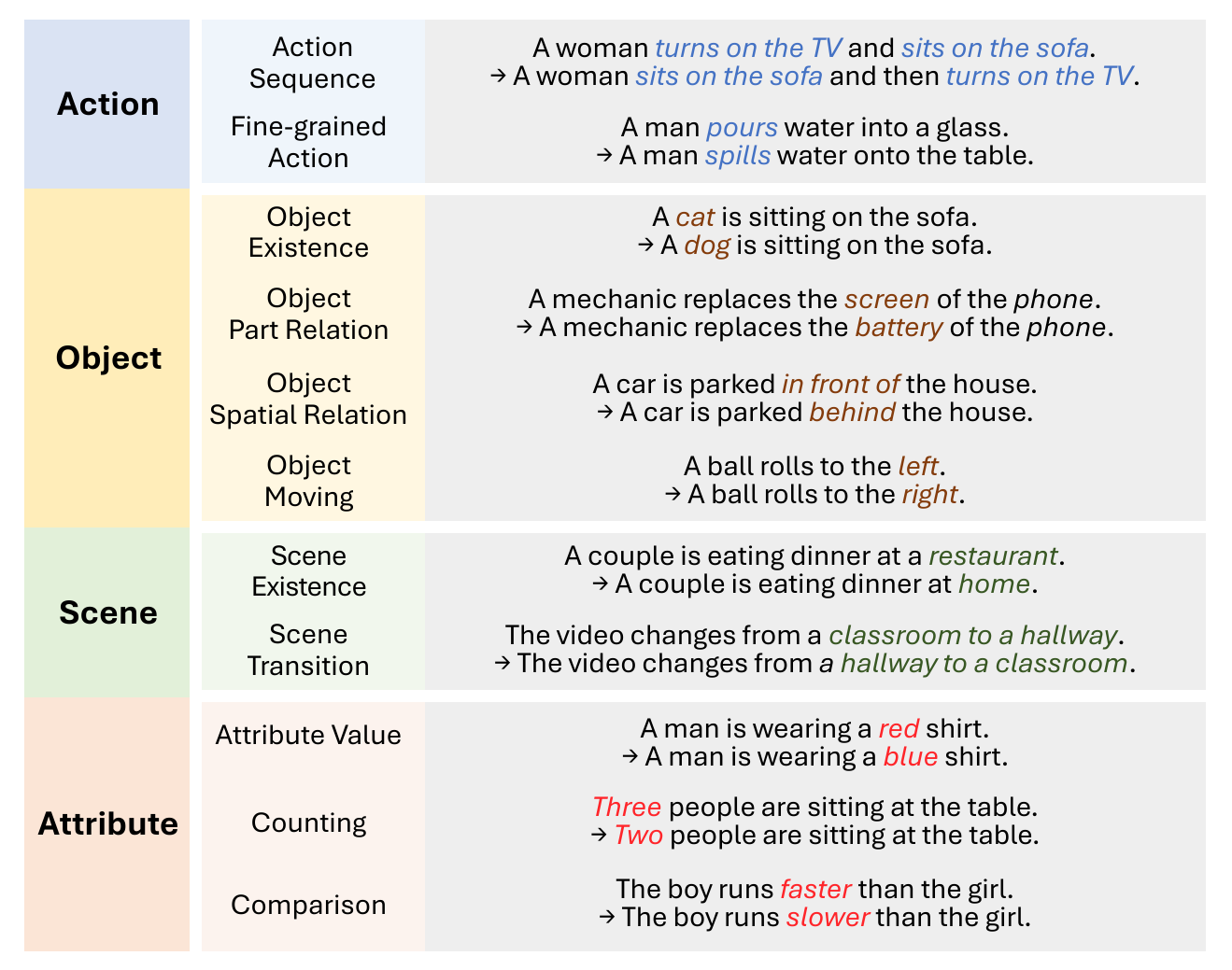}
\caption{Semantic relevance categories used in HI-VTG. The right column shows the original queries and modified queries according to each category.}
\label{fig_4}
\end{figure}

\paragraph{Explain Reward.}
Hard-irrelevant queries share high-level semantic context with the video but differ in specific details.
To correctly refuse hard-irrelevant queries, it is necessary to capture fine-grained semantic differences.
To encourage this capability, the explain reward $r_{\text{exp}}$ promotes refusal answers that explain the semantic mismatch for hard-irrelevant queries, and segment predictions for relevant queries.
\begin{equation}
r_{\mathrm{exp}}(o)
= \simf(a_{\text{pos}}, \hat{a}) - \simf(a_{\text{neg}}, \hat{a})
\label{eq:rar_compact}
\end{equation}
Here, $\hat{a}$ is the generated answer extracted from the \texttt{<answer>} section, and $\text{sim}(\cdot,\cdot)$ denotes the cosine similarity between SentenceBERT embeddings~\cite{sentencebert}.
For relevant queries, $a_{\text{time}}$ serves as the positive answer and $a_{\text{refusal}}$ as the negative.
For hard-irrelevant queries, $a_{\text{refusal}}$ serves as the positive answer and $a_{\text{time}}$ as the negative.
The reward encourages $\hat{a}$ to be closer to the positive reference and farther from the negative reference.
By encouraging answers that explicitly explain why the query does not correspond to the video in irrelevant cases, this reward enhances the model’s ability to capture fine-grained semantic differences, improving refusal performance for hard-irrelevant queries.

\paragraph{Query Correction Reward.}
To further enhance the understanding of fine-grained semantic differences between the video and an irrelevant query, we introduce a query correction reward.
This reward encourages the model to reconstruct the intended relevant query $q_{r}$, based on the video context $v$ and the given irrelevant query $q_{ir}$.
The correction reward is defined as:
\begin{equation}
r_{\text{cor}}(o) =
\begin{cases}
0, 
& q \in \text{Rel.},\\[4pt]
\text{sim}(q_{r}, \hat{c}), 
& q \in \text{Irre.}
\end{cases}
\label{eq_correct_reward}
\end{equation}
Here, $q_{r}$ denotes the original relevant query, and $\hat{c}$ is the corrected query extracted from the \texttt{<correct>} section of the generated output.
This reward is applied only when the input query is irrelevant, while it remains 0 for relevant queries since no correction is needed.
By reconstructing $q_{r}$ from $v$ and $q_{ir}$, the model performs semantic comparison between the video and the query, which enhances fine-grained reasoning and leads to more accurate refusal explanations for hard-irrelevant queries.

\subsection{Hard-Irrelevant VTG Dataset}
\label{sec_3_3}
We introduce a Hard-Irrelevant VTG (HI-VTG) dataset that includes hard-irrelevant queries and refusal answers.
Hard-irrelevant queries $q_{ir}$ are semantically close to the video but not actually relevant, and the refusal answers $a_{refusal}$ describe why the given query is not relevant to the video.

\Cref{fig_3} illustrates the overall construction process.
We first define semantic relevance categories that characterize possible relationships between queries and videos, and extract category types from the original query $q_{r}$.
Then, using the video descriptions, extracted categories, and the original query, we generate hard-irrelevant queries $q_{ir} $ and their refusal answers $a_{refusal}$.

We collect 2.5K videos from HowTo100M~\cite{howto100m}, YT-Temporal~\cite{yt-temporal}, DiDeMo~\cite{didemo}, QuerYD~\cite{queryd}, and InternVID~\cite{internvid}.
Text queries and timestamp annotations are obtained from existing VTG datasets, including VTG-IT~\cite{vtg-it}, TimeIT~\cite{timechat}, TimePro~\cite{timesuite}, LongVid~\cite{longvid}, and HTStep~\cite{htstep}.
Using these video–query pairs, we construct 7.5K hard-irrelevant queries and their corresponding refusal answers following the process described above.

\vspace{-0.1cm}
\paragraph{Semantic Relevance Category Extraction.}
To generate hard-irrelevant queries, we first define semantic relevance categories.
As shown in \cref{fig_4}, considering various properties of the video, we define 11 semantic relevance categories grouped into four high-level types (action, object, scene, and attribute).
These categories describe different possible relationships between the query and the video.
Then, we extract the semantic relevance categories of each original query.
We provide the original query and the definitions of the relevance categories to GPT-5-mini~\cite{gpt}, which then selects the top three categories that best characterize the relationship between the query and the video.
These selected categories are then used to modify the original query to generate a hard-irrelevant query.

\vspace{-0.1cm}
\paragraph{Hard-Irrelevant Query \& Refusal Answer Generation.}
After extracting the semantic relevance categories, we generate hard-irrelevant queries and their corresponding refusal answers.
Given the original query and the selected relevance categories, we modify one, two, or three semantic elements in the query, resulting in different degrees of semantic mismatch.
The degree of modification determines the irrelevance level: Strong Hard-Irrelevant (one element modified), Moderate Hard-Irrelevant (two elements modified), and Weak Hard-Irrelevant (three elements modified).
For each hard-irrelevant query, we prompt GPT-5-mini~\cite{gpt} to generate a refusal answer that explains why the query does not match the video.
The refusal answer explicitly states the mismatched semantic elements by comparing the query with the video description according to the selected relevance categories.

Using this procedure, we construct 10K query–answer pairs:
2.5K relevant pairs from the original VTG annotations and 7.5K Hard-Irrelevant pairs (Strong, Moderate, Weak) with their refusal explanations.
Our HI-VTG training dataset is used to fine-tune LVLMs under the RA-RFT strategy, enabling the model to learn both when to refuse and how to explain the refusal clearly.

\section{Experiments}

\begin{table*}[t]
\centering
\scriptsize
\setlength{\tabcolsep}{2pt}
\begin{tabular}{cl|ccccccc|ccccccc|ccccccc}
\toprule
\multirow{3}{*}{Type} & \multirow{3}{*}{Method} & 
\multicolumn{7}{c|}{HI-ActivityNet} &
\multicolumn{7}{c|}{HI-TVGBench} &
\multicolumn{7}{c}{HI-Charades}\\
\cline{3-23} \rule{0pt}{1.2EM}
& & \multicolumn{4}{c|}{RA-IoU} & \multicolumn{3}{c|}{F1 score} & \multicolumn{4}{c|}{RA-IoU} & \multicolumn{3}{c|}{F1 score} & \multicolumn{4}{c|}{RA-IoU} & \multicolumn{3}{c}{F1 score}\\
\cline{3-23} \rule{0pt}{1.2EM}
& & R@0.3 & R@0.5 & R@0.7 & mIoU & re. & irre. & avg. & R@0.3 & R@0.5 & R@0.7 & mIoU & re. & irre. & avg. & R@0.3 & R@0.5 & R@0.7 & mIoU & re. & irre. & avg. \\
\midrule
\multirow{2}{*}{VLP}
& RaTSG*~\cite{ratsg}  & 40.5 & 31.91 & 23.8 & 32.4 & 67.0 & 36.5 & 51.7 & 49.9 & 43.4 & 36.1 & 42.1 & 64.8 & 52.8 & 58.8 & - & - & - & - & - & - & -\\
& NA-VMR*~\cite{navmr} & - & - & - & - & - & - & - & 40.6 & 34.6 & 23.7 & 31.3 & 66.0 & 28.0 & 47.0 & - & - & - & - & - & - & - \\
\midrule
\multirow{3}{*}{SFT}
& TimeChat~\cite{timechat}   & 12.7 & 6.4 & 2.8 & 9.0 & 66.7 & 0.0 & 33.3 & 8.8 & 6.1 & 2.5 & 6.0 & 66.7 & 0.0 & 33.3 & 25.0 & 15.7 & 6.4 & 15.8 & 66.7 & 0.00 & 33.3  \\
& TimeSuite~\cite{timesuite} & 20.1 & 12.7 & 7.7 & 15.4 & 66.6 & 3.0 & 34.8 & 11.6 & 6.3 & 2.6 & 7.9 & 66.5 & 1.2 & 33.9 & 25.8 & 16.2 & 7.5 & 17.3 & 66.7 & 0.2 & 33.4 \\
& TRACE~\cite{trace}         & 18.2 & 12.8 & 8.3 & 14.1 & 65.9 & 10.3 & 38.1 & 11.4 & 8.4 & 6.3 & 9.0 & 64.2 & 13.3 & 38.7 & 18.0 & 10.1 & 5.1 & 12.0 & 66.2 & 4.2 & 35.2 \\
\midrule
\multirow{6}{*}{RFT}
& Time-R1~\cite{time-r1} & 53.0 & 45.0 & 38.2 & 45.9 & 74.2 & 66.9 & 70.5 & 42.2 & 36.8 & 31.6 & 37.3 & 69.9 & 59.0 & 64.5 & 59.1 & 50.9 & 40.0 & 48.9 & 72.8 & 63.7 & 68.2 \\
& \cellcolor{rowgray}+ RA-RFT (ours)  & \cellcolor{rowgray}59.6 & \cellcolor{rowgray}51.3 & \cellcolor{rowgray}43.8 & \cellcolor{rowgray}51.9 & \cellcolor{rowgray}77.6 & \cellcolor{rowgray}75.0 & \cellcolor{rowgray}76.3 & \cellcolor{rowgray}51.6 & \cellcolor{rowgray}46.3 & \cellcolor{rowgray}40.9 & \cellcolor{rowgray}46.6 & \cellcolor{rowgray}70.1 & \cellcolor{rowgray}69.9 & \cellcolor{rowgray}70.0 & \cellcolor{rowgray}62.4 & \cellcolor{rowgray}55.5 & \cellcolor{rowgray}45.8 & \cellcolor{rowgray}53.1 & \cellcolor{rowgray}72.5 & \cellcolor{rowgray}69.2 & \cellcolor{rowgray}70.8 \\
& VideoChat-R1~\cite{videochat-r1}  & 46.1 & 38.3 & 31.7 & 39.4 & 70.2 & 58.6 & 64.4 & 42.4 & 37.5 & 33.3 & 38.2 & 62.5 & 58.8 & 60.7 & 56.4 & 51.3 & 41.1 & 46.8 & 70.1 & 54.6 & 62.3 \\
& \cellcolor{rowgray}+ RA-RFT (ours)  & \cellcolor{rowgray}53.3 & \cellcolor{rowgray}45.0 & \cellcolor{rowgray}38.7 & \cellcolor{rowgray}46.5 & \cellcolor{rowgray}75.0 & \cellcolor{rowgray}70.8 & \cellcolor{rowgray}72.9 & \cellcolor{rowgray}46.4 & \cellcolor{rowgray}41.3 & \cellcolor{rowgray}36.4 & \cellcolor{rowgray}41.9 & \cellcolor{rowgray}68.9 & \cellcolor{rowgray}65.0 & \cellcolor{rowgray}66.9 & \cellcolor{rowgray}66.9 & \cellcolor{rowgray}62.3 & \cellcolor{rowgray}52.1 & \cellcolor{rowgray}57.6 & \cellcolor{rowgray}75.3 & \cellcolor{rowgray}71.3 & \cellcolor{rowgray}73.3\\
& VideoChat-R1-think~\cite{videochat-r1}  & 44.6 & 37.3 & 31.3 & 38.4 & 67.8 & 57.3 & 62.5 & 44.4 & 39.8 & 34.9 & 39.6 & 64.0 & 59.9 & 61.9 & 56.9 & 50.7 & 39.9 & 46.4 & 70.8 & 52.2 & 61.5\\
& \cellcolor{rowgray}+ RA-RFT (ours)  & \cellcolor{rowgray}52.6 & \cellcolor{rowgray}44.8 & \cellcolor{rowgray}38.6 & \cellcolor{rowgray}46.2 & \cellcolor{rowgray}73.4 & \cellcolor{rowgray}70.2 & \cellcolor{rowgray}71.8 & \cellcolor{rowgray}48.9 & \cellcolor{rowgray}44.3 & \cellcolor{rowgray}39.4 & \cellcolor{rowgray}44.0 & \cellcolor{rowgray}68.3 & \cellcolor{rowgray}67.1 & \cellcolor{rowgray}67.7 & \cellcolor{rowgray}66.1 & \cellcolor{rowgray}61.3 & \cellcolor{rowgray}50.2 & \cellcolor{rowgray}56.5 & \cellcolor{rowgray}74.8 & \cellcolor{rowgray}69.7 & \cellcolor{rowgray}72.3\\
\bottomrule
\end{tabular}
\vspace{-0.15cm}
\caption{Evaluation results on the Hard-Irrelevant VTG datasets. The methods marked with * are trained and evaluated under the same data distribution (in-distribution setting), while methods without * are evaluated in a zero-shot setting.}
\label{tab_1}
\end{table*}

\begin{table*}[t]
\centering
\scriptsize
\setlength{\tabcolsep}{2pt}
\begin{tabular}{cl|ccccccc|ccccccc|ccccccc}
\toprule
\multirow{3}{*}{Type} & \multirow{3}{*}{Method} & 
\multicolumn{7}{c|}{SS-ActivityNet~\cite{ratsg}} &
\multicolumn{7}{c|}{SS-Charades~\cite{ratsg,navmr}} &
\multicolumn{7}{c}{SS-QVHighlights~\cite{navmr}} \\
\cline{3-23} \rule{0pt}{1.2EM}
& &
\multicolumn{4}{c|}{RA-IoU} &
\multicolumn{3}{c|}{F1 score} &
\multicolumn{4}{c|}{RA-IoU} &
\multicolumn{3}{c|}{F1 score} &
\multicolumn{4}{c|}{RA-IoU} &
\multicolumn{3}{c}{F1 score} \\
\cline{3-23} \rule{0pt}{1.2EM}
& & R@0.3 & R@0.5 & R@0.7 & mIoU & re. & irre. & avg. 
& R@0.3 & R@0.5 & R@0.7 & mIoU & re. & irre. & avg. 
& R@0.3 & R@0.5 & R@0.7 & mIoU & re. & irre. & avg. \\
\midrule
\multirow{2}{*}{VLP}
& RaTSG*~\cite{ratsg}  & 68.6 & 60.4 & 52.7 & 60.9 & 83.9 & 83.9 & 83.9 & 57.1 & 61.6 & 68.7 & 63.7 & 71.2 & 77.8 & 74.5 & - & - & - & - & - & - & -\\
& NA-VMR*~\cite{navmr}& - & - & - & - & - & - &  & 42.0 & 48.1 & 56.5 & 51.2 & 76.4 & 66.2 & 71.3 & 85.1 & 80.2 & 73.4 & 77.1 & 92.6 & 93.2 & 92.9\\
\midrule
\multirow{2}{*}{SFT}
& TimeChat~\cite{timechat}       & 11.5 & 5.9 & 2.7 & 50.0 & 66.7 & 0.0 & 33.3 & 15.3 & 9.3 & 3.7 & 9.8 & 66.7 & 0.0 & 33.3 & 6.1 & 2.8 & 0.8 & 4.7 & 66.7 & 0.0 & 33.3\\
& TRACE~\cite{trace}            & 17.8 & 13.0 & 9.6 & 14.7 & 65.7 & 14.9 & 40.3 & 19.4 & 11.0 & 5.5 & 12.8 & 66.3 & 3.5 & 34.9 & 8.6 & 6.7 & 5.8 & 8.2 & 65.5 & 16.9 & 41.2\\
\midrule
\multirow{6}{*}{RFT}
& Time-R1~\cite{time-r1} & 65.2 & 57.3 & 50.0 & 58.0 & 84.8 & 83.6 & 83.9 & 66.4 & 59.0 & 49.7 & 57.3 & 80.4 & 78.4 & 79.4 & 77.7 & 72.0 & 64.1 & 69.2 & 91.3 & 90.7 & 91.0\\
& \cellcolor{rowgray}+ RA-RFT (ours)  & \cellcolor{rowgray}82.8 & \cellcolor{rowgray}77.8 & \cellcolor{rowgray}70.4 & \cellcolor{rowgray}74.9 & \cellcolor{rowgray}94.8 & \cellcolor{rowgray}95.2 & \cellcolor{rowgray}95.0 & \cellcolor{rowgray}69.8 & \cellcolor{rowgray}63.4 & \cellcolor{rowgray}54.0 & \cellcolor{rowgray}61.0 & \cellcolor{rowgray}80.9 & \cellcolor{rowgray}81.7 & \cellcolor{rowgray}81.3 & \cellcolor{rowgray}82.8 & \cellcolor{rowgray}77.8 & \cellcolor{rowgray}70.4 & \cellcolor{rowgray}74.9 & \cellcolor{rowgray}94.8 & \cellcolor{rowgray}95.2 & \cellcolor{rowgray}95.0\\
& VideoChat-R1~\cite{videochat-r1}  & 63.2 & 56.2 & 50.9 & 57.9 & 86.2 & 86.8 & 86.5 & 69.9 & 65.1 & 56.0 & 61.2 & 81.1 & 80.2 &  80.7 & 76.5 & 70.8 & 62.2 & 68.6 & 90.5 & 91.0 & 90.8\\
& \cellcolor{rowgray}+ RA-RFT (ours)  & \cellcolor{rowgray}65.1 & \cellcolor{rowgray}57.8 & \cellcolor{rowgray}52.3 & \cellcolor{rowgray}59.5 & \cellcolor{rowgray}87.6 & \cellcolor{rowgray}88.9 & \cellcolor{rowgray}88.3 & \cellcolor{rowgray}71.1 & \cellcolor{rowgray}66.4 & \cellcolor{rowgray}58.2 & \cellcolor{rowgray}63.1 & \cellcolor{rowgray}83.1 & \cellcolor{rowgray}83.5 & \cellcolor{rowgray}83.3 & \cellcolor{rowgray}80.3 & \cellcolor{rowgray}73.6 & \cellcolor{rowgray}65.0 & \cellcolor{rowgray}71.5 & \cellcolor{rowgray}92.4 & \cellcolor{rowgray}93.1 & \cellcolor{rowgray}92.8\\
& VideoChat-R1-think~\cite{videochat-r1}  & 62.1 & 55.2 & 50.2 & 56.9 & 82.3 & 84.0 & 83.2 & 69.9 & 64.1 & 55.1 & 60.6 & 81.4 & 78.8 & 80.1 & 72.4 & 66.5 & 60.4 & 65.5 & 89.4 & 89.9 & 89.6\\
& \cellcolor{rowgray}+ RA-RFT (ours)  & \cellcolor{rowgray}64.7 & \cellcolor{rowgray}57.4 & \cellcolor{rowgray}52.2 & \cellcolor{rowgray}59.2 & \cellcolor{rowgray}86.0 & \cellcolor{rowgray}88.0 & \cellcolor{rowgray}87.0 & \cellcolor{rowgray}71.3 & \cellcolor{rowgray}66.6 & \cellcolor{rowgray}57.6 & \cellcolor{rowgray}63.0 & \cellcolor{rowgray}83.5 & \cellcolor{rowgray}83.6 & \cellcolor{rowgray}83.6 & \cellcolor{rowgray}80.5 & \cellcolor{rowgray}73.8 & \cellcolor{rowgray}64.4 & \cellcolor{rowgray}71.5 & \cellcolor{rowgray}93.7 & \cellcolor{rowgray}94.2 & \cellcolor{rowgray}94.0\\
\bottomrule
\end{tabular}
\vspace{-0.15cm}
\caption{Evaluation results on existing Simply-Shuffled RA-VTG datasets. The methods marked with * are trained and evaluated under the same data distribution (in-distribution setting), while methods without * are evaluated in a zero-shot setting.}
\label{tab_2}
\end{table*}

\begin{table*}[t!]
\centering
\scriptsize
\setlength{\tabcolsep}{3pt}
\begin{tabular}{cl|cccc|ccc|ccc}
\toprule
\multirow{2}{*}{Type} & \multirow{2}{*}{Method} &
\multicolumn{4}{c|}{RA-IoU} &
\multicolumn{3}{c|}{F1 score} &
\multicolumn{3}{c}{Explanation Quality} \\
\cline{3-12} \rule{0pt}{1.2EM}
& & R@0.3 & R@0.5 & R@0.7 & mIoU & re. & irre. & avg. & RT-IoU & SBert score & LLM score\\
\midrule
\multirow{2}{*}{VLP} & RaTSG~\cite{ratsg} & 32.4 & 28.6 & 26.4 & 30.4 & 58.7 & 44.9 & 51.8 & - & - & - \\
& NA-VMR~\cite{navmr} & 37.0 & 33.0 & 30.0 & 34.4 & 52.7 & 52.3 & 52.5 & - & - & -\\
\midrule
\multirow{2}{*}{SFT} & TimeChat~\cite{timechat}    & 7.0 & 4.5 & 2.5 & 5.4 & 66.7 & 0.0 & 33.3 & 0.0 & 0.00 & 0.00 \\
& TRACE~\cite{trace}          & 16.0 & 12.0 & 7.5 & 12.8 & 66.7 & 0.0 & 33.0 & 0.0 & 0.00 & 0.00 \\
\midrule
\multirow{6}{*}{RFT} & Time-R1~\cite{time-r1} & 50.5 & 43.0 & 35.0 & 41.1 & 66.1 & 48.1 & 57.1 & 12.0 & 0.27 & 1.16 \\
& \cellcolor{rowgray}+ RA-RFT (ours) & \cellcolor{rowgray}63.0 & \cellcolor{rowgray}59.5 & \cellcolor{rowgray}49.0 & \cellcolor{rowgray}54.8 & \cellcolor{rowgray}74.0 & \cellcolor{rowgray}68.5 & \cellcolor{rowgray}71.2 & \cellcolor{rowgray}19.8 & \cellcolor{rowgray}0.47 & \cellcolor{rowgray}2.13\\
& VideoChat-R1~\cite{videochat-r1} & 47.0 & 40.0 & 33.5 & 37.9 & 65.4 & 39.7 & 52.5 & 8.3 & 0.22 & 0.76 \\
& \cellcolor{rowgray}+ RA-RFT (ours) & \cellcolor{rowgray}57.5 & \cellcolor{rowgray}50.0 & \cellcolor{rowgray}40.5 & \cellcolor{rowgray}48.2 & \cellcolor{rowgray}70.9 & \cellcolor{rowgray}63.3 & \cellcolor{rowgray}67.1 & \cellcolor{rowgray}15.8 & \cellcolor{rowgray}0.43 & \cellcolor{rowgray}1.64\\
& VideoChat-R1-think~\cite{videochat-r1}  & 44.5 & 39.0 & 31.5 & 36.9 & 65.9 & 45.5 & 55.7 & 7.3 & 0.25 & 1.00 \\
& \cellcolor{rowgray}+ RA-RFT (ours) & \cellcolor{rowgray}52.5 & \cellcolor{rowgray}45.5 & \cellcolor{rowgray}38.5 & \cellcolor{rowgray}43.7 & \cellcolor{rowgray}68.4 & \cellcolor{rowgray}59.4 & \cellcolor{rowgray}63.9 & \cellcolor{rowgray}14.8 & \cellcolor{rowgray}0.38 & \cellcolor{rowgray}1.51\\
\bottomrule
\end{tabular}
\vspace{-0.15cm}
\caption{Evaluation results on the Human-Annotated RA-VTG dataset.}
\label{tab_3}
\vspace{-0.2cm}
\end{table*}

\subsection{RA-VTG Benchmarks}
We evaluate our method under three relevance-aware VTG scenarios:
(1) Hard-Irrelevant VTG evaluation settings,
(2) Simply-Shuffled RA-VTG settings, and
(3) Human-Annotated RA-VTG settings.


\noindent\textbf{Hard-Irrelevant VTG Evaluation Datasets.}
We generate hard-irrelevant queries and their refusal responses using the same LLM-based generation procedure applied in training.
Based on existing VTG benchmarks~\cite{activitynet,charades,egonlq,tacos,hirest}, we construct the following hard-irrelevant relevance-aware evaluation datasets:
(1) \textit{HI-ActivityNet} consists of long-duration videos with 17K test video–query pairs,
(2) \textit{HI-TVGBench} integrates evaluation samples from multiple VTG datasets with 1.6K test video–query pairs, and
(3) \textit{HI-Charades} contains indoor human activity videos with 3.7K test video–query pairs.
All of these datasets contain relevant and hard-irrelevant queries in a 1:1 ratio.

\noindent\textbf{Simply-Shuffled RA-VTG Datasets.}
We also evaluate on previously proposed relevance-aware datasets~\cite{ratsg,navmr}, where irrelevant queries are obtained by simply shuffling queries across different videos.
Following prior settings~\cite{ratsg,navmr}, we experiment on (1) \textit{SS-ActivityNet}, (2) \textit{SS-Charades}, and (3) \textit{SS-QVHighlights}.
Each dataset maintains a 1:1 ratio between relevant and simply-shuffled irrelevant queries.

\noindent\textbf{Human-Annotated RA-VTG Dataset.}
To avoid potential bias from using the same construction procedure for both training and evaluation, we create a Human-Annotated RA-VTG dataset consisting of 100 relevant and 100 hard-irrelevant cases.
We sample 100 video samples from multiple source VTG datasets~\cite{charades, activitynet, hirest, tacos, coin}, and human annotator manually write hard-irrelevant queries and their corresponding refusal responses.

\subsection{Metrics}
To evaluate refusal-aware VTG models, we use RA-IoU and F1 score.
Following prior work~\cite{ratsg}, RA-IoU jointly evaluates relevance prediction and temporal localization.
If the query is relevant and the model outputs a timestamp, RA-IoU is measured as the IoU between the predicted and ground-truth segments.
If the query is irrelevant and the model does not generate a target segment, the score is set to 1; otherwise, 0.
The F1 score, defined as the harmonic mean of precision and recall, reflects how well the model balances detecting relevant queries while avoiding incorrect predictions on irrelevant ones.

We also assess the quality of the refusal explanation for irrelevant queries using RT-IoU, Sentence-BERT score, and LLM score.
We introduce RT-IoU, which compares the semantic relevance categories mentioned in the generated refusal answer with those in the reference answer.
The Sentence-BERT similarity score~\cite{sentencebert} measures semantic similarity between the generated and reference answers.
The LLM-based semantic consistency score uses GPT-5-mini~\cite{gpt} to evaluate whether the explanation correctly expresses why the query does not match the video.

\subsection{Experimental Setup}
\noindent\textbf{Baselines.}
We compare our method against prior training-based relevance-aware VTG approaches~\cite{ratsg,navmr} and LVLM-based VTG approaches.
For all LVLM-based VTG models, we include a prompt instruction that makes the model refuse irrelevant queries and explain the reason.
SFT-based VTG models~\cite{timechat,timesuite,trace} are trained to imitate instruction-formatted answers and suffer from catastrophic forgetting of generalization capabilities~\cite{mitigates, shenfeld2025rl}, leading them to predict target segments regardless of query relevance.
On the other hand, RFT-based models adjust their outputs to maximize reward signals, which enables them to better generalize refusal behavior when irrelevant queries are given.
Therefore, we apply our RA-RFT strategy to RFT-based models, including Time-R1~\cite{time-r1}, VideoChat-R1, and VideoChat-R1-Thinking~\cite{videochat-r1}.

\noindent\textbf{Implementation Details.}
We adopt 7B-scale backbones for all open-sourced RFT-based VTG baselines, including Time-R1, VideoChat-R1, and VideoChat-R1-thinking~\cite{time-r1,videochat-r1}.
To balance training efficiency and memory usage, we uniformly sample video frames at 2 FPS and resize each video so that the total pixel count is approximately 2.8M.
During RA-RFT, we post-train the model for 3 epochs with a batch size of 16 and use the final checkpoint for evaluation.
We do not fine-tune the model on any downstream benchmarks.
All experiments are conducted on 8×NVIDIA A100 GPUs with ZeRO-3~\cite{rajbhandari2020zero} optimization.

\subsection{Experimental Results}
\Cref{tab_1} presents the results on the hard-irrelevant VTG benchmarks.
Our RA-RFT consistently improves both relevance discrimination and relevance-aware temporal grounding across all baselines and all hard-irrelevant VTG datasets.
This indicates that our RA-RFT helps the model capture fine-grained semantic differences between the query and the video.

\Cref{tab_2} reports the performance on simply-shuffled RA-VTG settings, following prior work~\cite{ratsg,navmr}.
RA-RFT also improves all baselines across these datasets.
Although RA-RFT is designed to strengthen fine-grained semantic reasoning for hard-irrelevant queries, it also enhances the model’s ability to capture coarse-grained relevance differences.

To demonstrate our method in more natural scenarios, we evaluate on the human-annotated relevance-aware VTG dataset, which includes hard-irrelevant queries and reasoning-rich refusal answers written by annotators.
As shown in \cref{tab_3}, RA-RFT improves both refusal behavior and explanation quality across all baselines.
Higher RT-IoU and LLM scores indicate that the generated refusal explanations are more consistent with human-written explanations.
This confirms that RA-RFT improves both refusal behavior and refusal explanation quality.
Overall, these results show that RA-RFT consistently improves refusal behavior and fine-grained semantic reasoning, demonstrating its generalizability across relevance-aware VTG scenarios.

\begin{table}[t]
\centering
\footnotesize
\setlength{\tabcolsep}{2.2pt}
\begin{tabular}{l|ccc|ccc}
\toprule
\multirow{2}{*}{Method} &
\multicolumn{3}{c|}{RA-IoU} & \multicolumn{3}{c}{Explanation Quality }\\ \cline{2-7} \rule{0pt}{1.2EM}
& F1 & R@.5 & mIoU & RT-IoU & SBert. & LLM. \\
\midrule
Time-R1~\cite{time-r1} & 70.5 & 45.0 & 45.8 & 30.4 & 0.42 & 2.00\\
+ GRPO w/o HI-VTG & 70.0 & 46.1 & 46.4 & 28.3 & 0.38 & 1.85\\
+ GRPO w/ R.IoU & 75.1 & 51.2 & 50.9 & 34.5 & 0.47 & 2.29\\
+ GRPO w/ R.IoU-Exp & 75.6 & \textbf{51.6} & \textbf{52.2} & 36.0 & 0.48 & 2.37\\
+ GRPO w/ R.IoU-Exp-Cor & \textbf{76.3} & \underline{51.3} & \underline{51.9} & \textbf{37.1} & \textbf{0.51} & \textbf{2.44}\\
\bottomrule
\end{tabular}
\vspace{-0.15cm}
\caption{Ablation study on the reward components in RA-RFT. \textbf{Bold} and \underline{underlined} values indicate the best and second-best performances, respectively.}
\label{tab_4}
\end{table}

\begin{table}[t]
\centering
\footnotesize
\setlength{\tabcolsep}{2.0pt}
\begin{tabular}{l|cc|cc|cc}
\toprule
\multirow{2}{*}{Method} &
\multicolumn{2}{c|}{Strong Hard.} &
\multicolumn{2}{c|}{Moderate Hard.} & 
\multicolumn{2}{c}{Weak Hard.} \\
\cline{2-7} \rule{0pt}{1.2EM}
& F1 & LLM. & F1 & LLM. & F1 & LLM. \\
\midrule
Time-R1~\cite{time-r1} & 60.0 & 1.59 & 77.5 & 2.19 & 86.3 & 2.39 \\
+ GRPO w/o HI-VTG & 55.6 & 1.48 & 73.9 & 2.00 & 83.0 & 2.21 \\
+ GRPO w/ R.IoU & 68.0 & 1.95 & 83.7 & 2.42 & 91.8 & 2.62 \\
+ GRPO w/ R.IoU-Exp & 70.0 & 2.05 & 85.2 & 2.49 & \textbf{93.1} & \textbf{2.67} \\
+ GRPO w/ R.IoU-Exp-Cor & \textbf{70.6} & \textbf{2.09} & \textbf{86.8} & \textbf{2.63} & \textbf{93.1} & \textbf{2.67} \\
\bottomrule
\end{tabular}
\vspace{-0.15cm}
\caption{Ablation study across different levels of hard-irrelevance.}
\label{tab_5}
\vspace{-0.3cm}
\end{table}

\section{Analysis}
We analyze the effectiveness of our proposed
method and introduce its details. All ablation studies and discussions are conducted on the HI-ActivityNet dataset.
\subsection{Ablation Study}
We conduct ablation studies for reward components.
As shown in \cref{tab_4}, training the model using only simply-shuffled irrelevant queries with refuse-IoU reward does not help to refuse hard-irrelevant cases, since it focuses only on coarse relevance differences.
When trained with HI-VTG using only the refuse-IoU reward, the model shows improved relevance discrimination.
Adding the explain reward improves the performances by encouraging the model to state why the query does not match the video.
Finally, adding the query correction reward further improves F1 and explanation scores while maintaining RA-IoU performance, since reconstructing the original relevant query enhances fine-grained semantic understanding.
In addition, \cref{tab_5} shows that the performance gains are larger in strong hard-irrelevant cases.
This indicates that RA-RFT and the HI-VTG dataset effectively enhance the fine-grained semantic reasoning and refuse hard-irrelevant queries.

\begin{table}[t]
\centering
\footnotesize
\setlength{\tabcolsep}{2.0pt}
\begin{tabular}{l|cc|cc|cc}
\toprule
\multirow{2}{*}{Method} &
\multicolumn{2}{c|}{Strong Hard.} &
\multicolumn{2}{c|}{Moderate Hard.} & 
\multicolumn{2}{c}{Weak Hard.} \\
\cline{2-7} \rule{0pt}{1.2EM}
& F1 & LLM sc. & F1 & LLM sc. & F1 & LLM sc. \\
\midrule
RaTSG & 31.8 & - & 42.2 & - & 47.9 & - \\
\midrule
Time-R1 & 60.0 & 1.59 & 77.5 & 2.19 & 86.3 & 2.38 \\
+ \cellcolor{rowgray}RA-RFT (ours) & \cellcolor{rowgray}70.6 & \cellcolor{rowgray}2.09 & \cellcolor{rowgray}86.8 & \cellcolor{rowgray}2.63 & \cellcolor{rowgray}93.1 & \cellcolor{rowgray}2.67 \\
VideoChat-R1 & 53.0 & 1.30 & 67.9 & 1.83 & 78.8 & 2.20 \\
\cellcolor{rowgray}+ RA-RFT (ours) & \cellcolor{rowgray}65.9 & \cellcolor{rowgray}1.89 & \cellcolor{rowgray}82.9 & \cellcolor{rowgray}2.56 & \cellcolor{rowgray}89.7 & \cellcolor{rowgray}2.66\\
VideoChat-R1-think & 54.7 & 1.25 & 67.7 & 1.79 & 77.7 & 2.13 \\
\cellcolor{rowgray}+ RA-RFT (ours) & \cellcolor{rowgray}67.6 & \cellcolor{rowgray}1.88 & \cellcolor{rowgray}83.2 & \cellcolor{rowgray}2.52 & \cellcolor{rowgray}90.4 & \cellcolor{rowgray}2.71\\
\bottomrule
\end{tabular}
\vspace{-0.15cm}
\caption{Performance across different levels of hard-irrelevance.}
\label{tab_6}
\end{table}

\begin{table}[t]
\centering
\footnotesize
\begin{tabular}{l|cccc}
\toprule
\multirow{2}{*}{Method} &
\multicolumn{4}{c}{IoU score} \\
\cline{2-5} \rule{0pt}{1.2EM}
& R@.3 & R@.5 & R@.7 & mIoU \\
\midrule
RaTSG* & 53.1 & 38.8 & 22.5 & 39.8  \\
\midrule
Time-R1 & 59.9 & 40.1 & 22.0 & 41.2  \\
\cellcolor{rowgray}+ RA-RFT (ours) & \cellcolor{rowgray}60.3 & \cellcolor{rowgray}40.2 & \cellcolor{rowgray}21.7 & \cellcolor{rowgray}41.3  \\
VideoChat-R1 & 53.0 & 33.9 & 17.9 & 36.4  \\
\cellcolor{rowgray}+ RA-RFT (ours) & \cellcolor{rowgray}52.9 & \cellcolor{rowgray}32.9 & \cellcolor{rowgray}16.0 & \cellcolor{rowgray}35.5 \\
VideoChat-R1-think & 51.3 & 32.6 & 16.1 & 34.9 \\
\cellcolor{rowgray}+ RA-RFT (ours) & \cellcolor{rowgray}52.7 & \cellcolor{rowgray}32.8 & \cellcolor{rowgray}16.2 & \cellcolor{rowgray}35.4 \\
\bottomrule
\end{tabular}
\vspace{-0.15cm}
\caption{VTG performance on original VTG dataset. * indicates the model trained and evaluated under the same data distribution.}
\vspace{-0.3cm}
\label{tab_7}
\end{table}

\subsection{Discussions}
\label{sec_5_2}
\noindent\textbf{Performance Across Hard-Irrelevance Levels.}
\Cref{tab_6} shows the performance across different levels of hard-irrelevance. Our method consistently improves refusal behavior and explanation quality across all levels and all baselines, with larger gains observed in strong hard-irrelevant cases.
This demonstrates that our method strengthens the model’s fine-grained semantic reasoning, improving refusal performance on hard-irrelevant queries.

\noindent\textbf{Preserving Standard VTG Performance.}
\Cref{tab_7} reports VTG performance on the original VTG datasets. The grounding accuracy remains comparable before and after applying our method, indicating that enhancing refusal behavior does not degrade temporal grounding ability. In other words, our method improves relevance-aware reasoning while preserving the core temporal grounding performance of the VTG model.

\section{Conclusion}
We presented Refusal-Aware Reinforcement Fine-Tuning (RA-RFT) to effectively refuse hard-irrelevant queries in Video Temporal Grounding.
Built on the GRPO framework, RA-RFT integrates four complementary reward objectives—format, refuse-IoU, explain, and query correction—to enhance relevance discrimination and fine-grained semantic reasoning.
To support this, we introduced the Hard-Irrelevant VTG (HI-VTG) dataset containing hard-irrelevant queries and their refusal answers.
Experiments across multiple relevance-aware VTG scenarios show that RA-RFT improves refusal behavior and explanation quality while preserving standard grounding performance, demonstrating its effectiveness for robust and interpretable video-language reasoning.

\small \bibliographystyle{ieeenat_fullname} \bibliography{main}
\clearpage
\setcounter{page}{1}
\maketitlesupplementary
\setcounter{section}{0}
\renewcommand{\thesection}{\Alph{section}}

\section{Metrics}
We provide detailed descriptions of the evaluation metrics used in the main paper, including RA-IoU, F1 score, RT-IoU, Sentence-BERT score, and LLM-based score.

\noindent\textbf{RA-IoU.}
RA-IoU is an mIoU-based metric that incorporates relevance classification, following prior work~\cite{ratsg}.
(1) If the query is relevant and the model outputs a timestamp, we compute the IoU between the predicted segment and the ground-truth segment.
(2) If the query is irrelevant and the model does not produce a timestamp, we assign a score of 1.
(3) Otherwise, we assign a score of 0.
R@m denotes the proportion of samples whose RA-IoU is greater than a threshold $m$.

\noindent\textbf{RT-IoU.}
RT-IoU measures the alignment between the semantic relevance category mentioned in a refusal answer and the ground-truth category used to construct its corresponding hard-irrelevant query.
When extracting the categories from the refusal answer, we provide the generated refusal answer and the definitions of the semantic categories as input to GPT-5-mini.
Using the extracted categories and the ground-truth categories, RT-IoU is computed as the intersection of categories divided by the union of categories.


\noindent\textbf{SBert Score.}
We compute the cosine similarity between the Sentence-BERT~\cite{sentencebert} embeddings of the generated refusal answer and the ground-truth refusal answer, producing a score in the range of 0 to 1.

\noindent\textbf{LLM Score.}
We use GPT-5-mini~\cite{gpt} to assess the semantic consistency between the generated refusal answer and the ground-truth refusal answer, with the LLM producing a consistency score in the range of 1 to 5.

\section{Semantic Relevance Category Definition}
\begin{figure}[h]
\centering
\includegraphics[width=0.95\linewidth]{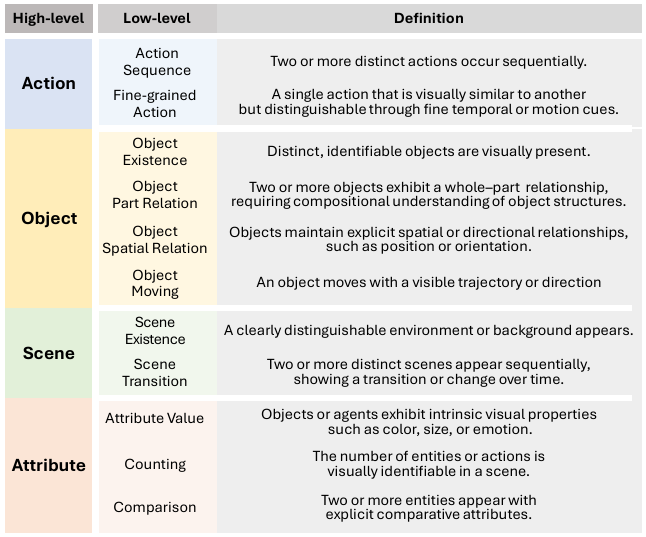}
\vspace{-0.15cm}
\caption{Semantic Relevance Category definition}
\label{category definition}
\vspace{-0.2cm}
\end{figure}

To construct hard-irrelevant queries, we define eleven semantic relevance categories grouped into four high-level types.
These categories describe the possible relationships between a video and a text query, and are defined to reflect the spatiotemporal characteristics of the video.
\cref{category definition} provides detailed descriptions of each semantic relevance category.


\section{Experimental Results via Category Types}
\begin{figure}[h]
\centering
\includegraphics[width=0.95\linewidth]{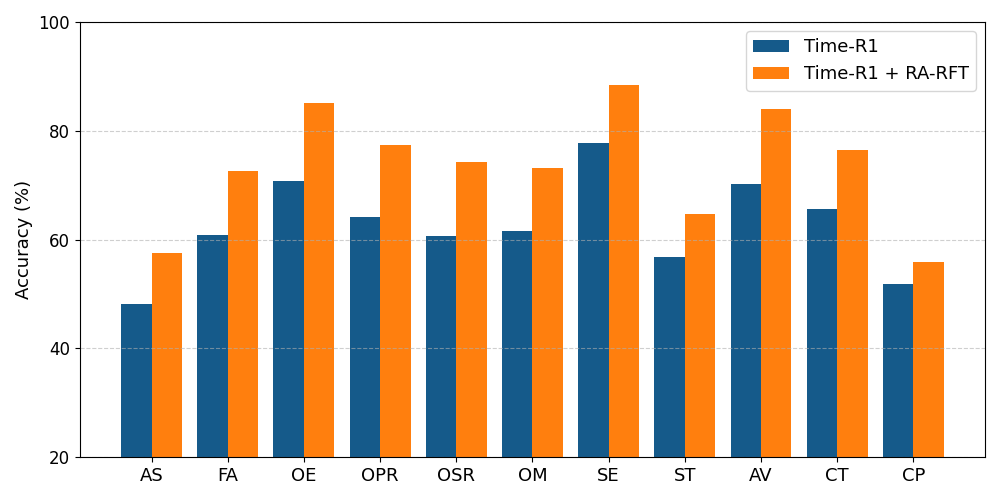}
\vspace{-0.15cm}
\caption{Performance analysis via semantic relevance categories.}
\label{performance via category}
\vspace{-0.15cm}
\end{figure}

To analyze relevance discrimination across semantic relevance categories, we evaluate performance using the categories employed to construct the hard-irrelevant queries.
\Cref{performance via category} shows that our model outperforms the baseline across all categories.
The model achieves strong improvements across categories requiring spatial understanding, such as Object Existence, Scene Existence, and Attribute Value, as well as categories involving temporal understanding, including Action Sequence, Object Moving, and Scene Transition. These results indicate that our approach effectively enhances relevance discrimination across a wide range of semantic mismatch types.


\section{Comparison of Supervised Fine-Tuning Method}
To analyze the effectiveness of the proposed RA-RFT strategy, we compare our method with supervised fine-tuning (SFT). \Cref{tab_abla_finetuning} shows the results of training on our HI-VTG dataset using SFT. The SFT-trained model shows reduced performance across all metrics, likely due to catastrophic forgetting caused by imitating instruction-formatted answers. In contrast, the model trained with RA-RFT achieves higher RA-IoU and F1 scores, demonstrating the effectiveness of our approach.


\vspace{-0.1cm}
\begin{table}[h]
\centering
\footnotesize
\setlength{\tabcolsep}{3pt}
\begin{tabular}{c|cccc|ccc}
\toprule
\multirow{2}{*}{Method} &
\multicolumn{4}{c|}{RA-IoU} & \multicolumn{3}{c}{F1-Score}\\
\cline{2-8} \rule{0pt}{1.2EM}
& R@.3 & R@.5 & R@.7 & mIoU & rel. & irrel. & avg. \\
\midrule
Time-R1 & 53.0 & 45.0 & 38.2 & 45.9 & 74.2 & 66.9 & 70.5 \\
+SFT & 29.5 & 20.4 & 11.8 & 20.9 & 0.1 & 0.0 & 0.0 \\
\cellcolor{rowgray}+RA-RFT (ours)& \cellcolor{rowgray}59.6 & \cellcolor{rowgray}51.3 & \cellcolor{rowgray}43.8 & \cellcolor{rowgray}51.9 & \cellcolor{rowgray}77.6 & \cellcolor{rowgray}75.0 & \cellcolor{rowgray}76.3 \\
\bottomrule
\end{tabular}
\vspace{-0.2cm}
\caption{Comparison with other fine-tuning methods.}
\label{tab_abla_finetuning}
\end{table}

\begin{table*}[t]
\centering
\scriptsize
\setlength{\tabcolsep}{2pt}
\begin{tabular}{cl|ccc|ccc|ccc}
\toprule
\multirow{2}{*}{Type} & \multirow{2}{*}{Method} &
\multicolumn{3}{c|}{RA-ActivityNet} &
\multicolumn{3}{c|}{RA-Charades} &
\multicolumn{3}{c}{RA-TVGBench} \\
\cline{3-5} \cline{6-8} \cline{9-11} \rule{0pt}{1.2EM}
& & RT-IoU & SBert score & LLM score & RT-IoU & SBert score & LLM score & RT-IoU & SBert score & LLM score\\
\midrule
\multirow{3}{*}{SFT} & TimeChat~\cite{timechat} & 0.0 & 0.00 & 0.00 & 0.0 & 0.00 & 0.00 & 0.0 & 0.00 & 0.00\\
& TimeSuite~\cite{timesuite} & 0.0 & 0.00 & 0.00 & 0.0 & 0.00 & 0.00 & 0.0 & 0.00 & 0.00 \\
& TRACE~\cite{trace}                  & 0.0 & 0.00 & 0.01 & 0.0 & 0.00 & 0.00 & 0.0 & 0.02 & 0.00\\
\midrule
\multirow{6}{*}{RFT} & Time-R1~\cite{time-r1} & 30.4 & 0.41 & 2.00 & 31.7 & 0.40 & 1.76 & 24.3 & 0.38 & 1.51\\
& \cellcolor{rowgray}+ RA-RFT (ours) & \cellcolor{rowgray}37.1 & \cellcolor{rowgray}0.51 & \cellcolor{rowgray}2.44 & \cellcolor{rowgray}39.1 & \cellcolor{rowgray}0.49 & \cellcolor{rowgray}2.10 & \cellcolor{rowgray}33.7 & \cellcolor{rowgray}0.55 & \cellcolor{rowgray}2.18 \\
& VideoChat-R1~\cite{videochat-r1} & 25.9 & 0.35 & 1.72 & 25.3 & 0.32 & 1.44 & 18.4 & 0.29 & 1.28 \\
& \cellcolor{rowgray}+ RA-RFT (ours) & \cellcolor{rowgray}34.4 & \cellcolor{rowgray}0.48 & \cellcolor{rowgray}2.32 & \cellcolor{rowgray}40.7 & \cellcolor{rowgray}0.49 & \cellcolor{rowgray}2.33 & \cellcolor{rowgray}26.6 & \cellcolor{rowgray}0.41 & \cellcolor{rowgray}1.85 \\
& VideoChat-R1-think~\cite{videochat-r1} & 24.6 & 0.34 & 1.67 & 20.9 & 0.28 & 1.28 & 27.8 & 0.42 & 1.88\\
& \cellcolor{rowgray}+ RA-RFT (ours) & \cellcolor{rowgray}34.3 & \cellcolor{rowgray}0.48 & \cellcolor{rowgray}2.32 & \cellcolor{rowgray}37.7 & \cellcolor{rowgray}0.46 & \cellcolor{rowgray}2.23 & \cellcolor{rowgray}30.6 & \cellcolor{rowgray}0.45 & \cellcolor{rowgray}1.99 \\
\bottomrule
\end{tabular}
\vspace{-0.1cm}
\caption{Refusal explanation quality on the RA-VTG evaluation datasets.}
\label{tab_refusal_explanation}
\vspace{-0.15cm}
\end{table*}



\section{Refusal Explanation Quality on the HI-VTG Dataset}
\Cref{tab_refusal_explanation} presents the refusal explanation quality on the three RA-VTG evaluation datasets. Models trained with RA-RFT achieve higher scores across RT-IoU, SBERT score, and LLM-based score compared to the base models. These results indicate that our method improves the model’s ability to generate clearer and more appropriate refusal explanations for hard-irrelevant queries across diverse evaluation settings.

\section{More Experimental Results via Difficulty}
To complement the main paper’s analysis in \cref{sec_5_2}, we further evaluate RA-RFT across different levels of hard-irrelevance on other datasets. \Cref{tab_sup_hicharades} and \cref{tab_sup_hitvgbench} show the HI-Charades and HI-VTGBench results. RA-RFT consistently improves both refusal accuracy and explanation quality at all difficulty levels, with the largest gains appearing in the strong hard-irrelevant setting where fine-grained reasoning is crucial. Overall, these extended results across multiple datasets show that RA-RFT generalizes well across different levels of semantic discrepancy and consistently improves both refusal ability and explanation quality in diverse hard-irrelevant VTG scenarios.
\begin{table}[h]
\centering
\footnotesize
\setlength{\tabcolsep}{2.0pt}
\begin{tabular}{l|cc|cc|cc}
\toprule
\multirow{2}{*}{Method} &
\multicolumn{2}{c|}{Strong Hard.} &
\multicolumn{2}{c|}{Moderate Hard.} & 
\multicolumn{2}{c}{Weak Hard.} \\
\cline{2-7} \rule{0pt}{1.2EM}
& F1 & LLM sc. & F1 & LLM sc. & F1 & LLM sc. \\
\midrule
RaTSG & 53.6 & - & 66.3 & - & 74.5 & - \\
\midrule
Time-R1 & 63.2 & 1.68 & 73.3 & 1.76 & 82.8 & 1.98  \\
+ \cellcolor{rowgray}RA-RFT (ours) & \cellcolor{rowgray}71.5 & \cellcolor{rowgray}1.95 & \cellcolor{rowgray}82.1 & \cellcolor{rowgray}2.22 & \cellcolor{rowgray}90.0 & \cellcolor{rowgray}2.29 \\
VideoChat-R1 & 50.7 & 1.19 & 65.8 & 1.61 & 73.3 & 1.77  \\
\cellcolor{rowgray}+ RA-RFT (ours) & \cellcolor{rowgray}70.4 & \cellcolor{rowgray}2.09 & \cellcolor{rowgray}84.6 & \cellcolor{rowgray}2.57 & \cellcolor{rowgray}90.3 & \cellcolor{rowgray}2.54\\
VideoChat-R1-think & 47.7 & 1.08 & 59.3 & 1.37 & 72.1 & 2.48  \\
\cellcolor{rowgray}+ RA-RFT (ours) & \cellcolor{rowgray}69.1 & \cellcolor{rowgray}2.00 & \cellcolor{rowgray}82.1 & \cellcolor{rowgray}2.43 & \cellcolor{rowgray}88.5 & \cellcolor{rowgray}2.48\\
\bottomrule
\end{tabular}
\vspace{-0.1cm}
\caption{Performance across different levels of hard-irrelevance on HI-Charades dataset.}
\label{tab_sup_hicharades}
\end{table}
\begin{table}[h]
\centering
\footnotesize
\setlength{\tabcolsep}{2.0pt}
\begin{tabular}{l|cc|cc|cc}
\toprule
\multirow{2}{*}{Method} &
\multicolumn{2}{c|}{Strong Hard.} &
\multicolumn{2}{c|}{Moderate Hard.} & 
\multicolumn{2}{c}{Weak Hard.} \\
\cline{2-7} \rule{0pt}{1.2EM}
& F1 & LLM sc. & F1 & LLM sc. & F1 & LLM sc. \\
\midrule
Time-R1 & 52.5 & 0.95 & 66.2 & 1.49 & 78.8 & 2.08  \\
+ \cellcolor{rowgray}RA-RFT (ours) & \cellcolor{rowgray}67.5 & \cellcolor{rowgray}1.57 & \cellcolor{rowgray}84.7 & \cellcolor{rowgray}2.20 & \cellcolor{rowgray}91.6 & \cellcolor{rowgray}2.77 \\
VideoChat-R1 & 58.0 & 0.80 & 74.6 & 1.29 & 80.7 & 1.74  \\
\cellcolor{rowgray}+ RA-RFT (ours) & \cellcolor{rowgray}64.1 & \cellcolor{rowgray}1.41 & \cellcolor{rowgray}76.1 & \cellcolor{rowgray}1.82 & \cellcolor{rowgray}85.6 & \cellcolor{rowgray}2.30\\
VideoChat-R1-think & 62.4 & 1.36 & 76.4 & 2.05 & 77.1 & 2.23  \\
\cellcolor{rowgray}+ RA-RFT (ours) & \cellcolor{rowgray}70.1 & \cellcolor{rowgray}1.62 & \cellcolor{rowgray}81.6 & \cellcolor{rowgray}2.03 & \cellcolor{rowgray}85.6 & \cellcolor{rowgray}2.33 \\
\bottomrule
\end{tabular}
\vspace{-0.1cm}
\caption{Performance across different levels of hard-irrelevance on HI-TVGBench dataset.}
\label{tab_sup_hitvgbench}
\vspace{-0.3cm}
\end{table}

\section{Performance Analysis of the Qwen2.5-VL-7B Base Model Trained From Scratch}
To demonstrate the effectiveness of our HI-VTG training data and RA-RFT strategy on a general LVLM, we apply our method to the Qwen2.5-VL-7B model, which is not originally trained for the VTG task. The model is trained for 3 epochs.
As shown in \cref{tab_scratch}, with our method incorporated, the model achieves higher performance in both VTG grounding and relevance discrimination.
This indicates that the our dataset and learning strategy are effective in enabling the model to refuse irrelevant queries and perform temporal grounding.


\begin{table}[h]
\centering
\footnotesize
\setlength{\tabcolsep}{3pt}
\begin{tabular}{c|cccc|ccc}
\toprule
\multirow{2}{*}{Method} &
\multicolumn{4}{c|}{RA-IoU} & \multicolumn{3}{c}{F1-Score}\\
\cline{2-8} \rule{0pt}{1.2EM}
& R@.3 & R@.5 & R@.7 & mIoU & rel. & irrel. & avg. \\
\midrule
Qwen2.5-VL-7B & 32.9 & 26.2 & 21.2 & 27.8 & 69.8 & 41.2 & 55.5 \\
\cellcolor{rowgray}+RA-RFT (ours) & \cellcolor{rowgray}55.4 & \cellcolor{rowgray}48.6 & \cellcolor{rowgray}42.9 & \cellcolor{rowgray}49.6 & \cellcolor{rowgray}75.2 & \cellcolor{rowgray}73.1 & \cellcolor{rowgray}74.2 \\
\bottomrule
\end{tabular}
\vspace{-0.1cm}
\caption{Performance of Qwen2.5-VL-7B trained from scratch.}
\label{tab_scratch}
\end{table}
\vspace{-0.1cm}

\begin{figure*}[t]
\centering
\includegraphics[width=0.9\linewidth]{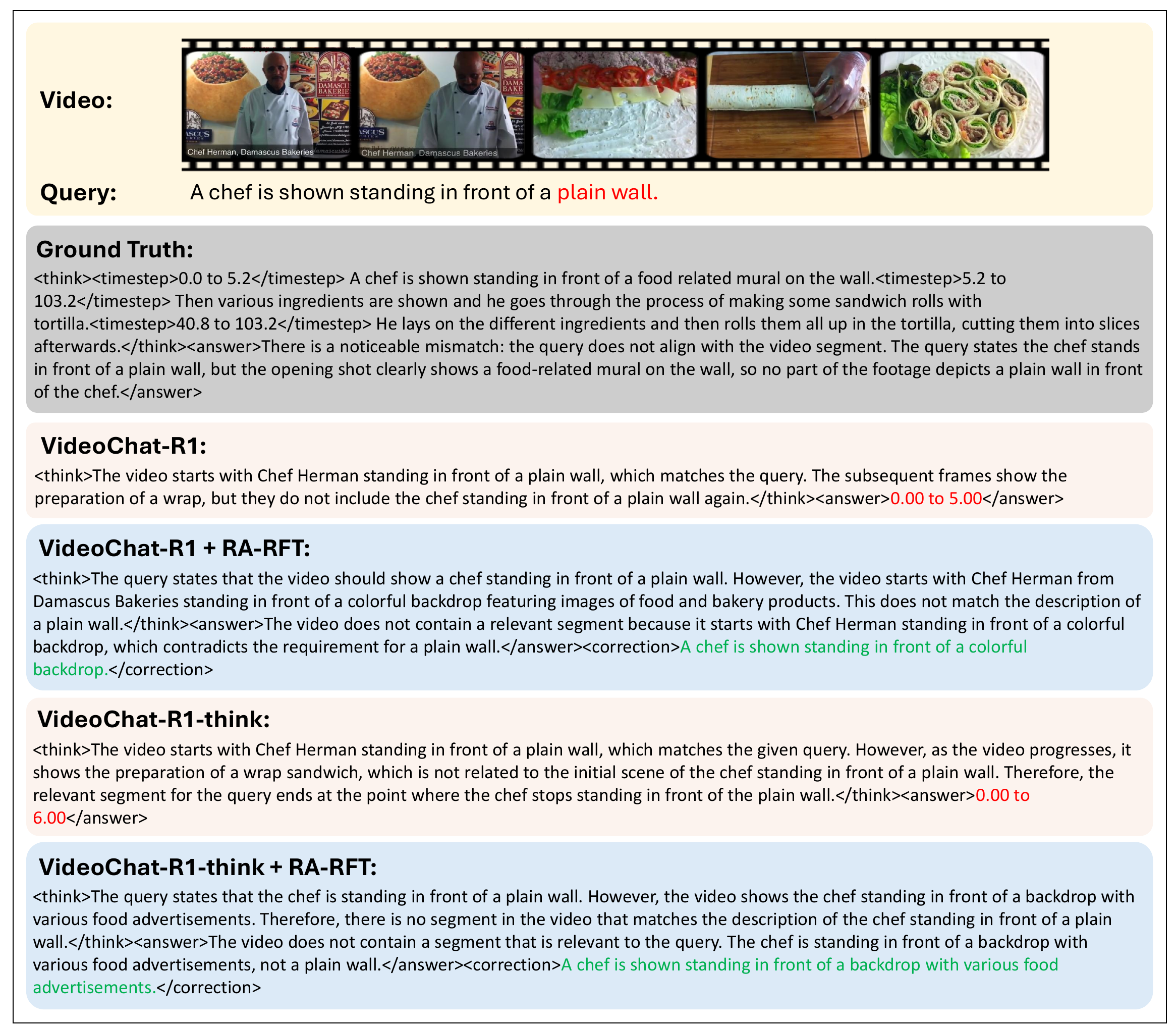}
\vspace{-0.1cm}
\caption{Qualitative results for strong hard-irrelevant queries from HI-ActivityNet.}
\label{fig_qua_1}
\end{figure*}
\vspace{-0.1cm}

\begin{figure*}[t]
\centering
\includegraphics[width=0.9\linewidth]{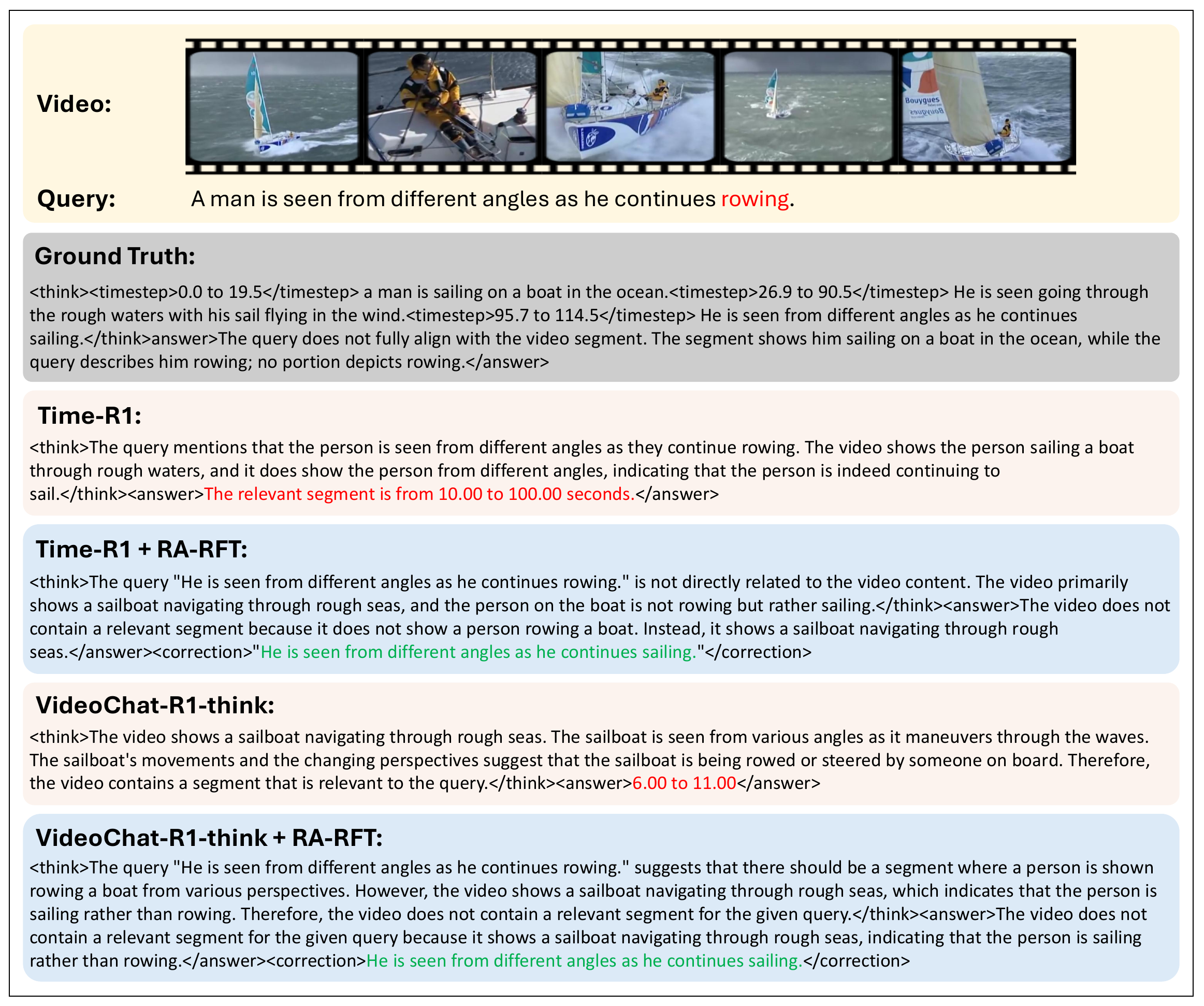}
\vspace{-0.1cm}
\caption{Additional qualitative results for strong hard-irrelevant queries from HI-ActivityNet.}
\label{fig_qua_2}
\end{figure*}
\vspace{-0.15cm}

\begin{figure*}[t]
\centering
\includegraphics[width=0.9\linewidth]{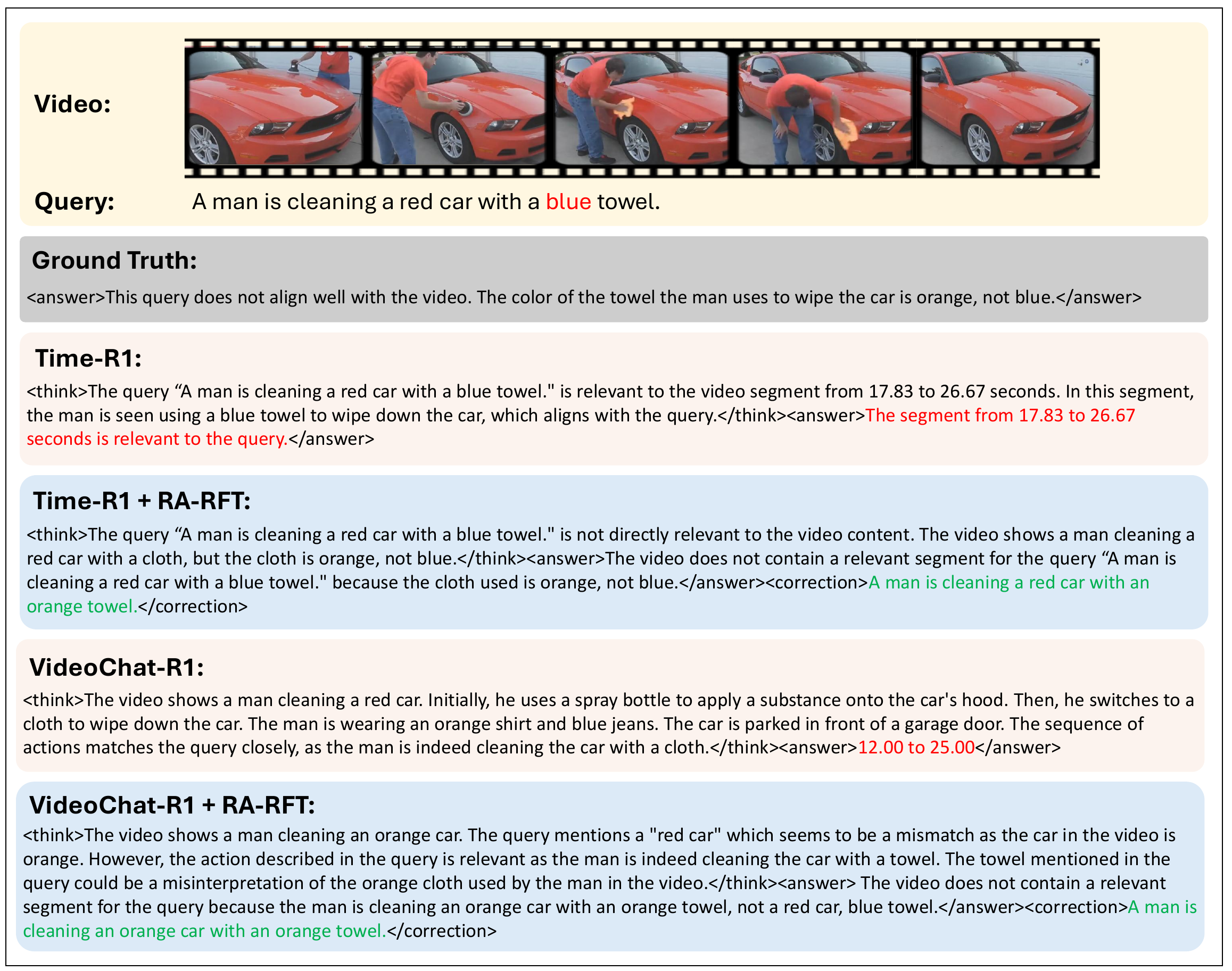}
\vspace{-0.1cm}
\caption{Qualitative results for hard-irrelevant queries from human-annotated RA-VTG dataset.}
\label{fig_qua_3}
\end{figure*}
\vspace{-0.15cm}

\section{Qualitative Results}
\Cref{fig_qua_1} and \cref{fig_qua_2} present qualitative results for strong hard-irrelevant queries from HI-ActivityNet. Also, \Cref{fig_qua_3} shows qualitative results for hard-irrelevant queries from the human-annotated RA-VTG dataset. While previous methods often fail to refuse hard-irrelevant queries and predict temporal segments, the models trained with RA-RFT effectively refuse these hard-irrelevant queries. In addition, our models clearly explain the refusal reasons and correctly reconstruct the original queries.

\begin{table*}[t]
\centering
\aboverulesep=0ex 
\belowrulesep=0ex 
\footnotesize
\setlength{\tabcolsep}{0.45em}     
\begin{tabular}{|l|}
\midrule \rule{0pt}{1.0EM}
[SYSTEM]
You are a multi-label classifier for building Video Temporal Grounding (VTG) datasets. \\
\\
\#\# Task introduction \\
- The input “related\_query” is a valid VTG query. \\
- Do NOT rewrite the query. \\
- Identify category paths that can modify the query into a plausible but video-irrelevant (negative) query. \\
- Select all applicable category paths and briefly justify why each can produce an irrelevant query. \\
\\
\#\# Category taxonomy (Parent / Child) \\
\\
\#\#\# Action \\
- Action/ActionSequence — Change the temporal ordering of actions. \\
- Action/FineGrainedAction — Replace an action verb with a visually similar but directionally or temporally distinct one. \\
\\
\#\#\# Object \\
- Object/ObjectExistence — Add or remove an identifiable object. \\
- Object/ObjectPartRelation — Modify part–whole relations or accessory relations. \\
- Object/ObjectSpatialRelation — Change relative spatial positions of objects. \\
- Object/ObjectMoving — Change the motion direction or trajectory of an object. \\
\\
\#\#\# Scene \\
- Scene/SceneExistence — Replace the type of scene. \\
- Scene/SceneTransition — Change scene order, transition direction, or timing. \\
\\
\#\#\# Attribute \\
- Attribute/AttributeValue — Change intrinsic properties such as color, size, material, shape, or state. \\
- Attribute/Counting — Change the number of objects or actions. \\
- Attribute/Comparison — Flip comparative relations such as size or speed. \\
\\
\#\# Output requirements \\
- Output ONLY the JSON below. \\
- Use exact "Parent/Child" names. \\
- Return all applicable, distinct categories, sorted by diagnostic strength. \\
- Output at least 3 categories whenever possible. \\
\\
\textnormal{[}OUTPUT JSON\textnormal{]} \\
\{ \\
  \quad “eligible\_categories”: [ \\
    \quad\quad \{“path”: “Parent/Child”, “reason”: “\textless justification\textgreater”\}, \\
    \quad\quad \{“path”: “Parent/Child”, “reason”: “\textless justification\textgreater”\} \\
  \quad ] \\
\} \\
\midrule
\end{tabular}
\caption{Prompt for extracting semantic relevance categories from a given query using an LLM.}
\label{prompt_1}
\end{table*}

\begin{table*}[t]
\centering
\aboverulesep=0ex 
\belowrulesep=0ex 
\footnotesize
\setlength{\tabcolsep}{0.45em}     
\begin{tabular}{|l|}
\midrule \rule{0pt}{1.0EM}
[SYSTEM] 
You generate hard negative (irrelevant) queries for Video Temporal Grounding (VTG).\\
\\
\#\# Purpose\\
- Create irrelevant (negative) queries from valid related queries.\\
- For each negative query, generate structured reasoning that a Video-LLM could output.\\
\\
\#\# Task\\
- Input: related\_query, reference timestamp, optional video\_context, and category-based plans.\\
- For each plan:\\
\quad 1. Edit the related\_query ONLY along its categories to produce one irrelevant query.\\
\quad 2. Generate reasoning using the REQUIRED block format:\\
\quad\quad \textless irrelevant\_answer\textgreater...\textless /irrelevant\_answer\textgreater
\textless category1\textgreater...\textless /category1\textgreater
\textless category2\textgreater...\textless /category2\textgreater
...\\
\\
- \textless irrelevant\_answer\textgreater block:\\
\quad * Must state misalignment between the query and the video.\\
\quad * Strength depends on difficulty (high/medium/easy).\\
\\
- \textless category\textgreater  blocks:\\
\quad * One block per applied category (tag = path lowercased with slashes as underscores).\\
\quad * Briefly explain why the irrelevant query does not match the video for that category.\\
\\
- Difficulty levels:\\
\quad - 1 category → strong\\
\quad - 2 categories → moderated\\
\quad - 3 categories → weak\\
\\
\#\# Category taxonomy (Parent / Child) \\
(Same category taxonomy as in the semantic relevance category extraction prompt.)
\\\\
\#\# Input format \\
- related\_query: “\textless string\textgreater”\\
- related\_query\_timestamp: “\textless start\textgreater-\textless end\textgreater second”\\
- plans: list of 1–3 items, each with difficulty and applied\_categories.\\
- video\_context:\\
\quad * A textual description of the video, provided as a string OR a JSON array of strings.\\
\quad * Each entry may include an associated time range and a natural-language description of what appears in the video.\\
\\
\#\# Output format (JSON only)\\
The model must output a single JSON object.  
Each difficulty present in the input plans must appear once under “negs”.\\
All fields must be included exactly as shown.\\
\\
\{\\
\quad “negs”: \{\\
\quad\quad "\textless difficulty\textgreater”: \{\\
\quad\quad\quad “irrel\_query”: “\textless generated negative query\textgreater”, \quad\qquad\textit{\% negative query produced via category edits}\\
\quad\quad\quad “applied\_categories”: [\\
\quad\quad\quad\quad \{“path": “Parent/Child”\}, ... \quad\quad\quad\quad\quad\quad\quad\qquad\textit{\% same paths \& order as in the plan}\\
\quad\quad\quad ],\\
\quad\quad\quad “reasoning”:\\
\quad\quad\quad\quad “\textless irrelevant\_answer\textgreater ... \textless/irrelevant\_answer\textgreater”\\
\quad\quad\quad\quad “\textless parent\_child\textgreater ... \textless/parent\_child\textgreater” ... , \quad\quad\quad\enspace\textit{\% one block per applied category}\\
\quad\quad\quad “difficulty\_tag”: “\textless difficulty\textgreater” \quad\quad\quad\quad\quad\quad\quad\quad\enspace\textit{\% must match the plan}\\
\quad\quad \}\\
\quad \}\\
\}\\

\midrule
\end{tabular}
\caption{Prompt for generating hard-irrelevant queries and refusal answers. A given query, extracted semantic relevance categories, and video context are used to generate an irrelevant query using an LLM.}
\label{prompt_2}
\end{table*}

\begin{table*}[t]
\centering
\aboverulesep=0ex 
\belowrulesep=0ex 
\footnotesize
\setlength{\tabcolsep}{0.45em}     
\begin{tabular}{|p{0.75\textwidth}|}
\midrule \rule{0pt}{1.0EM}
[SYSTEM]\\
You are a strict multi-label classifier.\\
Identify which reasoning categories from the video–text mismatch categories below are used or implied in \\a Generated Response that explains why a query is irrelevant to a video.\\
Select all applicable categories according to the meaning expressed in the response.\\
\\
\#\# Video–Text Mismatch Categories\\
(Same category taxonomy as in the semantic relevance category extraction prompt.)\\
\\
\#\# Rules\\
1. Include a category only if it is clearly supported or implied by the reasoning.  \\
2. Multiple categories may apply, but avoid redundant or speculative labels.  \\
3. Use only the exact category paths listed above.  \\
4. Ignore style, tone, or fluency — focus purely on reasoning content.\\  
5. If none apply, return an empty list.\\
\\
\#\# Output Format\\
Return only a JSON array of strings containing the selected categories.\\
Examples:\\
\textnormal{[} “Object/ObjectExistence”, “Attribute/Counting”\textnormal{]}\\
If none apply: []\\
\midrule
\end{tabular}
\caption{Prompt for evaluating RT-IoU between the semantic categories in the model’s refusal answer and the ground-truth categories.}
\label{prompt_3}
\end{table*}

\begin{table*}[t]
\centering
\aboverulesep=0ex 
\belowrulesep=0ex 
\footnotesize
\setlength{\tabcolsep}{0.45em}     
\begin{tabular}{|p{0.75\textwidth}|}
\midrule \rule{0pt}{1.0EM}
[SYSTEM]\\
You are a evaluator designed to assess the reasoning consistency between a Generated Response and a Ground Truth (GT) Response.
\\\\
\#\# TASK:\\
Your job is to evaluate how faithfully the Generated Response reproduces the reasoning in the GT Response.\\
\\
\#\# INSTRUCTIONS:
\\
\#\#\# Reasoning Consistency Evaluation:\\
- Evaluate how faithfully the Generated Response reproduces the reasoning and justification in the GT Response.\\
- A consistent response must keep the GT’s mismatch points, evidence, and contextual explanations. It must not distort their meaning.\\
- Omissions or contradictions of GT reasoning elements must be penalized.\\
- Extra explanations are allowed if they are logically consistent with the GT.\\
- The reasoning must remain factually and logically compatible with the GT Response.\\
- Do not consider fluency, tone, or paraphrasing style. Focus only on semantic and factual consistency.\\
\\
\#\#\# Scoring Scale (0–5):\\
Assign a score between 0 and 5, allowing decimal values, based on how well the reasoning aligns with the ground-truth reasoning.\\

\\
\#\#\# Evaluation Mindset:\\
- You MUST prioritize factual and logical alignment over stylistic similarity.\\
- Do NOT penalize harmless elaborations.\\
- You MUST penalize any omission or contradiction of GT reasoning.\\
- You MUST NOT assign a score above 4.9 unless reasoning is perfectly consistent.\\
\\
\#\# OUTPUT:\\
Return ONLY a Python dictionary literal. No explanations.\\\\
Examples:\\
{`score': 4.0}\\
{`score': 1.5}\\
{`score': 3.7}\\
\midrule
\end{tabular}
\caption{Prompt for evaluating an LLM score between a model’s refusal answer and the ground-truth response.}
\label{prompt_4}
\end{table*}

\section{Prompt Details}
We used GPT-5-mini to construct the HI-VTG dataset.
\Cref{prompt_1} shows the prompts used for extracting semantic relevance categories from a given video and query. Also, \cref{prompt_2} shows the prompts used for generating hard-irrelevant queries and corresponding refusal answers.
To evaluate the explanation quality of the model outputs, we used the prompts in \Cref{prompt_3} and \cref{prompt_4}, and extracted RT-IoU and LLM-based scores.

\end{document}